\def\year{2022}\relax
\documentclass[letterpaper]{article} 
\usepackage{aaai23}  
\usepackage{times}  
\usepackage{helvet}  
\usepackage{courier}  
\usepackage[hyphens]{url}  
\usepackage{graphicx} 
\usepackage{natbib}  
\usepackage{caption} 
\DeclareCaptionStyle{ruled}{labelfont=normalfont,labelsep=colon,strut=off} 
\usepackage{algorithm}
\usepackage{algorithmic}

\usepackage{newfloat}
\usepackage{listings}
\floatstyle{ruled}
\newfloat{listing}{tb}{lst}{}
\floatname{listing}{Listing}
\usepackage{multirow}
\usepackage{booktabs}
\usepackage{subcaption}
\usepackage{makecell}
\usepackage{amsmath}
\usepackage{hhline}
\usepackage{boldline}
\DeclareMathOperator*{\argmax}{arg\,max}
\usepackage{pifont}
\newcommand{\cmark}{\ding{51}}%
\newcommand{\xmark}{\ding{55}}%
\usepackage{siunitx}
\usepackage{comment}
\usepackage{tabularx}
\usepackage{amsfonts,amssymb}

\title{3D Human Pose Lifting with Grid Convolution}
\author{
	Yangyuxuan Kang,\textsuperscript{\rm 1,\rm 2}\thanks{This work was done when they were interns at Intel Labs China, supervised by Anbang Yao.} Yuyang Liu,\textsuperscript{\rm 3$\ast$} Anbang Yao,\textsuperscript{\rm 4\dag} Shandong Wang,\textsuperscript{\rm 4} Enhua Wu\textsuperscript{\rm 1,\rm 2,\rm 5}\thanks{Corresponding authors.}
}
\affiliations{
	\textsuperscript{\rm 1} State Key Laboratory of Computer Science, Institute of Software, Chinese Academy of Sciences\\
	\textsuperscript{\rm 2} University of Chinese Academy of Sciences 
	\textsuperscript{\rm 3} Tsinghua University\\
	\textsuperscript{\rm 4} Intel Labs China 
	\textsuperscript{\rm 5} Faculty of Science and Technology, University of Macau\\
	kyyx@ios.ac.cn, yyliu22@mails.tsinghua.edu.cn, \{anbang.yao,shandong.wang\}@intel.com, ehwu@um.edu.mo
}

\begin{document}

\maketitle

\begin{abstract}
Existing lifting networks for regressing 3D human poses from 2D single-view poses are typically constructed with linear layers based on graph-structured representation learning. In sharp contrast to them, this paper presents Grid Convolution (GridConv), mimicking the wisdom of regular convolution operations in image space. GridConv is based on a novel Semantic Grid Transformation (SGT) which leverages a binary assignment matrix to map the irregular graph-structured human pose onto a regular weave-like grid pose representation joint by joint, enabling layer-wise feature learning with GridConv operations. We provide two ways to implement SGT, including handcrafted and learnable designs. Surprisingly, both designs turn out to achieve promising results and the learnable one is better, demonstrating the great potential of this new lifting representation learning formulation. To improve the ability of GridConv to encode contextual cues, we introduce an attention module over the convolutional kernel, making grid convolution operations input-dependent, spatial-aware and grid-specific. We show that our fully convolutional grid lifting network outperforms state-of-the-art methods with noticeable margins under (1) conventional evaluation on Human3.6M and (2) cross-evaluation on MPI-INF-3DHP. Code is available at https://github.com/OSVAI/GridConv.
\end{abstract}
\newcommand{\etal}{\textit{et al.}}
\newcommand{\wrt}{w.r.t}
\newcommand{\etc}{\textit{etc}}
\newcommand{\eg}{\textit{e.g.}}

\section{Introduction}
\label{sec:intro}
3D human pose estimation is essential for various applications.
The task aims to recover the 3D positions of human body joints from images or videos. Benefiting from great advances in deep learning techniques, 3D human pose estimation with a single image input has now become practical.

One mainstream solution family estimates 3D human pose in two stages. The first stage detects 2D pose in an image, and the second stage lifts detected 2D pose to its 3D estimate.
Along with the advent of many well-designed 2D pose detectors, such as HRNet~\cite{sun2019hrnet}, 2D human pose detection technology is gradually becoming mature, showing significantly improved performance, even in outdoor scenarios with dramatic changes of background and rarely seen situations with diverse occlusions.
Driven by this as well as the prevalence of effective methods to generate large amounts of 2D-to-3D human pose pairs, 2D-to-3D pose lifting has become a critical research topic, and thus has attracted increased attention recently. Many works~\cite{fang2018posegrammar,zhao2019semgcn,kang2020explicit} have been devoted to advancing 2D-to-3D pose lifting research.
These works typically represent human pose as a 1D feature vector or a \textit{graph}, and
use either fully connected network or graph convolutional network to regress 3D pose from 2D input.

However, we observe that a pretty successful family of deep learning techniques, convolutional neural networks for image recognition and editing tasks, does not attract the interest of researchers in the lifting field. A vital reason is that graph-structured human skeleton pose having unbalanced joint neighborhoods hinders the use of convolution operation with regular kernels.
Motivated by the observation, we address the pose lifting problem by formulating a novel grid-based representation learning paradigm, attempting to introduce a 2D coordinate system to measure joint relationships and enable regular convolution operations and advanced design of building blocks.
Regarding our goal, three critical questions need to be considered: (1) Is it possible to transform a human skeleton pose into an image-like grid coordinate system?
(2) How to preserve intrinsic joint relationships of human skeleton during the transformation?
(3) After such transformation, can we use few modifications to convolutional networks to pursue a high-performance lifting model?

To the first question, we propose \textit{Semantic Grid Transformation} (SGT) which maps the irregular graph-structured human pose onto a regular \textit{weave-like grid pose representation} joint by joint.
To the second question, we design a handcrafted layout that meanwhile preserves skeleton topology and brings in a new kind of motion semantics.
In addition, to explore a better grid layout, we propose a learning-based SGT called \textit{AutoGrids} that automatically searches the layout conditioned on the input distribution. 
To the last question, SGT enables a new type of standard convolution operations on the grid pose. We call this operation paradigm \textit{Grid Convolution} (GridConv).
We further enhance the learning capability of GridConv by introducing an attention module over the convolutional kernels, making GridConv input-dependent, spatial-aware and grid-specific.

The solutions for the above three questions constitute our core contributions to constructing a new category of fully convolutional network that lifts 2D pose to 3D estimate in the weave-like grid pose domain. Extensive experiments on public 3D human pose estimation datasets demonstrate superior performance of our fully convolutional lifting network to existing methods by using the proposed representation learning paradigm. Furthermore, our method retains its effectiveness in the augmented training regime with synthetic data or joint optimization of the 2D human pose detector and the 2D-to-3D lifting network, showing further improved performance.

\section{Related Work}
\subsection{End-to-end 3D Human Pose Estimation}
In the pre-deep-learning era, 3D human pose estimation usually relies on building a 2D pictorial model from images and inferring plausible 3D targets from 2D evidence by Bayesian probabilistic models~\cite{belagiannis20143dpictorial,andriluka2010tracking}.
With the booming of deep learning techniques and the availability of high-quality 3D human dataset in the community, tremendous progress has been achieved by end-to-end learning of 3D human pose estimation~\cite{pavlakos2017coarsetofine,mehta2017vnect,zhou2017towards,zhou2021hemlets}.
These approaches show significant advantages over traditional ones. 

With the rise of convolutional neural network techniques, the technology for a related task, namely 2D human pose detection, has become more and more mature. In recent years, many prevailing 2D human pose detectors, such as OpenPose~\cite{cao2019openpose} and HRNet~\cite{sun2019hrnet}, have been proposed. Under this context, one critical research problem arises: it is possible to infer 3D human pose directly from 2D pose detection? 
To this problem, an early work~\cite{zhou2016sparseness} used sparse representation on 3D pose and inferred 3D pose under the condition of giving 2D pose probability heatmap or coordinate. Later on, \cite{martinez2017simple} proposed the two-stage 3D human pose estimation paradigm with deep learning, which first detects 2D body keypoints in an image and then regresses 3D pose coordinate from 2D pose coordinate. Since then, a lot of methods have been proposed to improve 2D-to-3D pose lifting scheme.

\subsection{2D-to-3D Pose Lifting}
The pioneering lifting work \cite{martinez2017simple} treated input 2D pose as a generic 1D feature vector and directly regressed 3D joint coordinate.
Subsequent works attempted to leverage prior knowledge of the human body skeleton to improve lifting optimization.
For example, \cite{sun2017compositional} exploited joint connection structure by representing pose as a composition of bones.
\cite{Dabral2018strcutureandmotion} introduced illegal articulation angle penalty and body symmetry constraint in the training process.
\cite{fang2018posegrammar} modeled skeleton motion in three high-level aspects including kinematics, symmetry, and motor coordination by defining joint relations in a recurrent network.
Our work is similar to them in modeling high-level joint relations, but the proposed representation learning paradigm is differentiated from the others.

With the arising research of Graph Convolutional Networks (GCNs), many works
represented human pose as a graph by mapping joints and limbs as graph nodes and edges, and substituted fully connected networks by GCNs as their lifting models.
Most of them~\cite{ci2019lcn,zhao2019semgcn,cai2019exploitingspatialtemporal,liu2020comprehensivegcn,zou2020highorder} focused on developing pose-relevant graph convolution operators and network architectures.
Some works \cite{zeng2021skeletalgcn,hu2021conditionalgraph} argued that the default skeletal graph is sub-optimal for perceiving long-distance joint relations, and thus proposed to dynamically adjust the graph structure.

This work goes beyond graph representation for human pose and formulates a semantically more informative lifting representation learning paradigm.
Moreover, with the help of some sophisticated strategies that generate large-scale synthetic 2D-to-3D data~\cite{gong2021poseaug}, our method can get further improvement after fine-tuning with augmented data, showing its great generalization ability.

\begin{figure*}[htb]
	\centering
	\includegraphics[width=0.9\linewidth]{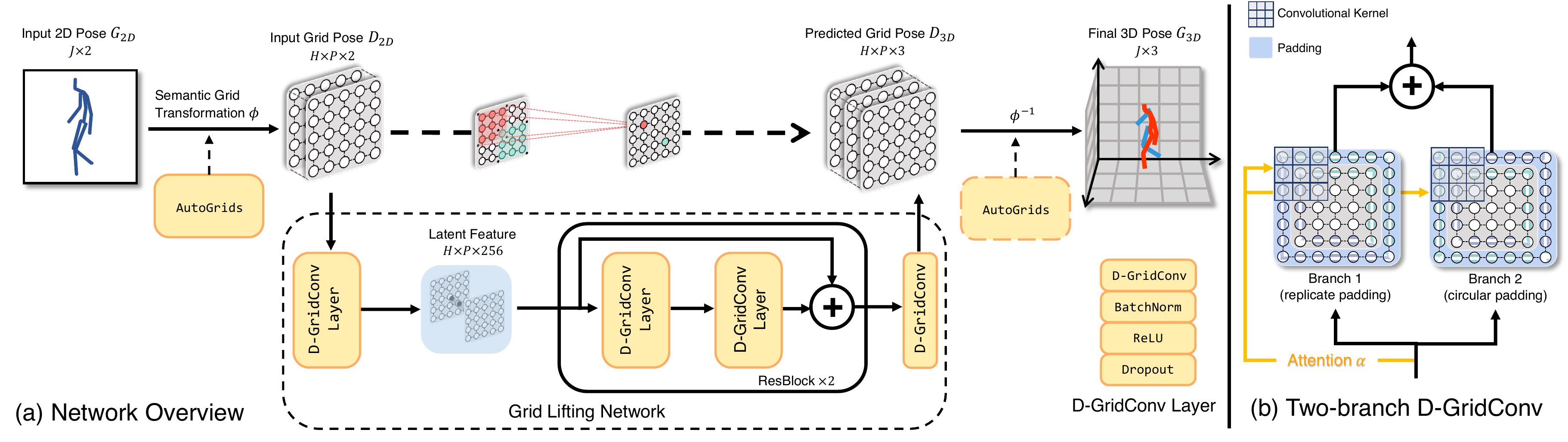}
	\caption{(a) \textbf{An architectural overview of the proposed grid lifting network.} Here, 2D grid pose $D_{_{2D}}$, namely the network input, is transformed from 2D human pose input $G_{_{2D}}$ via an SGT module $\phi$. D-GridConv layers learn latent feature embedding on the grid pose. At the end of the network, an inverse SGT module $\phi^{-1}$ rearranges 3D grid pose $D_{_{3D}}$ to target 3D human pose $G_{_{3D}}$. (b) \textbf{The internal architecture of D-GridConv module.} The output grid pose is obtained by summing up two-branched convolution results of padded inputs.}
	\label{fig:framework}
	\vspace{-1.5em}
\end{figure*}

\section{Method}
In this section, we first describe the formulation of Semantic Grid Transformation (SGT), which maps graph-structured human pose to a uniform weave-like grid pose representation, giving birth to Grid Convolution (GridConv). 
For SGT, we propose a handcrafted design and a learnable one.
Then we present the concept of GridConv as well as its dynamic form. 
Finally, we detail the architecture of our grid lifting network shown in Figure~\ref{fig:framework}.

\subsection{Semantic Grid Transformation}
\label{sec:sgt}
Suppose that we have a human pose $G\in \mathbb{R}^{J\times C}$, where $J$ denotes the number of body joints, $C$ denotes the coordinate dimensions for each joint ($C=2$ for 2D pose, and $C=3$ for 3D pose). The basic goal of SGT is to construct a \textit{grid pose} $D\in \mathbb{R}^{H\times P\times C}$, defined as a regular weave-like grid representation with the spatial size of $H$\texttimes $P$ filled by J body joints, where $HP$$\geq$$J$. By changing the setting of $H$\texttimes $P$, a grid pose $D$ could be either square or rectangular in shape, e.g., 5\texttimes5 or 7\texttimes3. SGT mapping function $\phi$ is defined as:
\begin{equation}
	D = \phi(G) = S \times G,
\end{equation}
where $S\in \{0,1\}^{HP\times J}$ is a binary assignment matrix which maps the graph-structured human pose $G$ to the desired grid pose $D$ joint by joint. 
During the mapping, a grid node $D_{{p}}, p\in [1, HP]$ in the grid pose $D$ will be filled by a particular body joint $G_j, j\in [1,J]$ only if $S_{{p},j}=1$.
Inverse SGT $\phi^{-1}$ that maps $D$ to $G$ is formed by inversing the assignment process, as referred to our supplementary material.

Recall that existing lifting methods typically adopt skeleton graph as pose representation. 
When constructing a grid pose $D$, it is natural to allow the desired grid pose to inherit joint features and preserve skeleton topology. Concretely, given an edge set of skeleton graph $\boldsymbol{E}$, this goal can be accomplished by adding the following two constraints to $\phi$:
\begin{eqnarray}
	\label{eqn_equivalence}
	S_{{p},i}\times \sum_{{q}\in\boldsymbol{N}({p})} S_{{q},j} \geq 1, \exists{p}\in[1,HP], &\forall (i,j)\in \boldsymbol{E} \\
	\label{eqn_unique}
	\sum_{k=1}^J S_{{p},k}=1, &\forall p\in[1,HP],
\end{eqnarray}
where $\boldsymbol{N}(\cdot)$ denotes the neighborhood of a certain grid node, namely four adjacent nodes in the horizontal and vertical directions.

On the one hand, by satisfying Equation~(\ref{eqn_equivalence}), originally connected body joints remain adjacent in the resulting grid pose. On the other hand,  Equation~(\ref{eqn_unique}) restricts each row of $S$ as a one-hot vector, which allows each grid node to have explicit semantic meaning (coordinate of a specific joint). 
Equation (\ref{eqn_equivalence}) and (\ref{eqn_unique}) produce replicants of some joints in the resulting grid layout and provide a loose collection of solutions.
This formulation of SGT earns two merits for incubating handcrafted design and the learnable one.

\noindent \textbf{Merits of SGT.} 
(1) Grid nodes having both vertical and horizontal edges offer multiple connection types for depicting joint relationships, which allows us to handcraft a semantically richer pose structure.
(2) A large number of solutions existing in the assignment space make it possible to define a learnable SGT by first describing the space in continuous distribution and then searching an optimal point.

\subsection{Two Designs for Implementing SGT}

In light of the above discussion, 
we define and analyze the advantages of SGT. 
Next, we provide a handcrafted SGT as a basic design.
And then, we present AutoGrids that automatically learns SGT as a generalized design.

\begin{figure}
	\centering
	\includegraphics[width=0.6\linewidth]{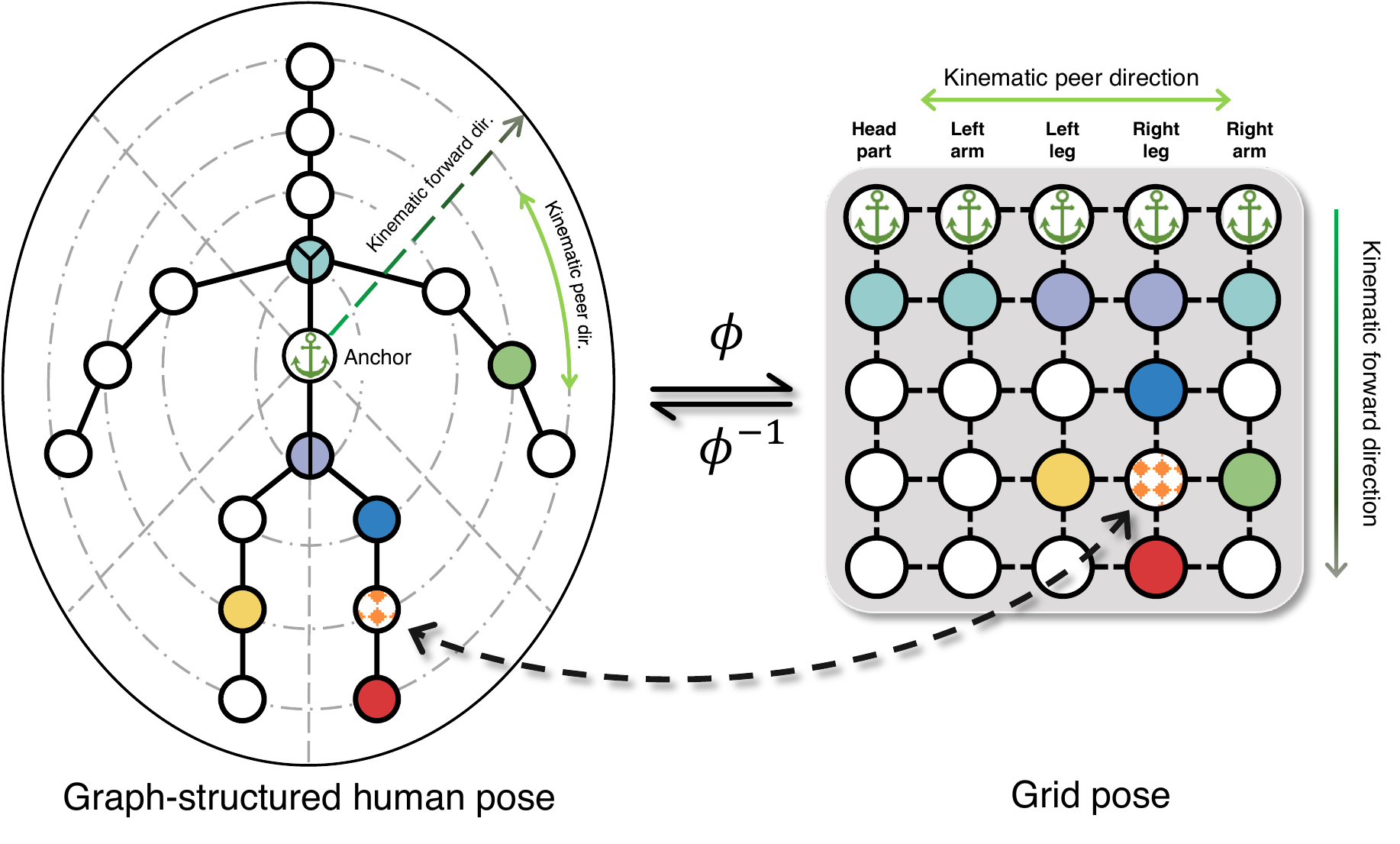}
	\caption{Handcrafted SGT. In a heuristic manner, torso joint is set as the anchor. The remaining joints are arranged along kinematic forward and peer directions into the vertical and horizontal directions of the grid structure. 
	}
	\label{fig:manual_prior}
	\vspace{-1.5em}
\end{figure}

\noindent \textbf{Handcrafted SGT design.} We heuristically make the resulting grid pose well encode both the vertical (along kinematic forward direction) and the horizontal (along kinematic peer direction) relationships of body joints to the root joint (e.g., torso joint), preserving prior joint connections of the skeleton pose graph structure. 
The corresponding grid layout is shown in Figure~\ref{fig:manual_prior}. Some joints have replicants in the grid, which are averaged during inverse SGT.
In Section~4, we test the efficacy of such a handcrafted layout and compare it with many other variants. 
This well addresses the first two questions we raised in Introduction: \textit{whether grid representation is feasible (answer: SGT); and how to preserve prior knowledge in skeleton graph (answer: handcrafted SGT).}

Although handcrafted SGT already achieves remarkable performance,  it still faces some issues, such as the scenario using a new definition of skeleton, where redesigning of SGT is required.
It motivates us to seek an automatic formulation and to further excavate the learning potential of grid representation.

\noindent \textbf{Learnable SGT design.}  
We shelve the constraint on preserving prior graph-structured joint connections defined in Equation~(\ref{eqn_equivalence}), and propose a learnable module called \textit{AutoGrids} to learn an adaptive assignment matrix conditioned on the input human skeleton pose, which is jointly optimized with our lifting network (its architecture will be clarified later).

\begin{figure}[t]
	\centering
	\includegraphics[width=0.6\linewidth]{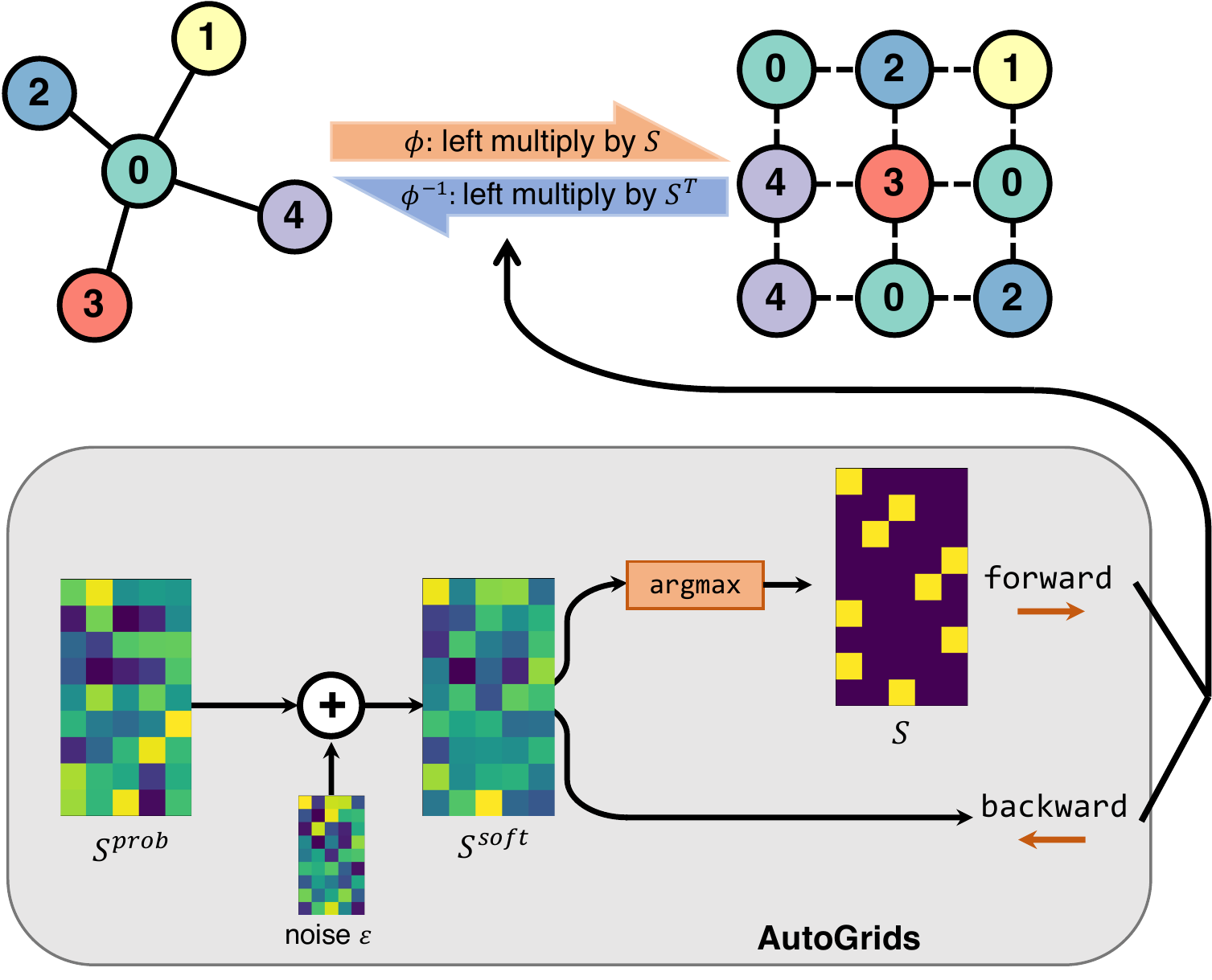}
	\caption{Illustration of the learning process of AutoGrids.}
	\label{fig:gat}
	\vspace{-1.5em}
\end{figure}

To learn an assignment matrix $S$ filled by discrete binary values, the difficulty lies in how to make the training process differentiable. To address this problem, we adopt Gumbel Softmax~\cite{jang2016gumbelsoftmax} which uses a continuous distribution of assignment matrix to approximate the sampling of $S$. Let $S^{prob}\in \mathbb{R}^{HP\times J}$ be a probability distribution of an assignment matrix filled by continuous positive values, whose element $S^{prob}_{ij}$ indicates the probability score assigning joint $G_j$ of skeleton pose graph to grid node $D_i$.

During training, AutoGrids module generates a soft assignment matrix $S^{soft}$ by:
\begin{equation}
	S^{soft} = S^{prob} + \varepsilon,
\end{equation}
where $\varepsilon\in \mathbb{R}^{HP\times J}$ is a Gumbel noise that assists to resample the soft assignment matrix $S^{soft}$ from the probability distribution $S^{prob}$. 
For forward inference, the desired binary assignment matrix $S$ can be easily determined by taking the highest probability response per row on $S^{soft}$ according to:
\begin{equation}
	S_i = onehot\big(\argmax_j\, S^{soft}_{ij}\big).
\end{equation}
This discretization operation cuts off the backward gradient flow during training, so we use straight-through estimator~\cite{courbariaux2015binaryconnect_straightthrough} for parameter update. Specifically, in the backward, continuous gradient approximation is used to directly update $S^{soft}$. 

Introduced noise interference encourages the exploration on different grid pose proposals, which facilitates AutoGrids module to identify a decent grid layout.
Figure~\ref{fig:gat} illustrates the learning process of AutoGrids. In the implementation, AutoGrids module is jointly trained with the lifting network in an end-to-end manner. Experiments in Section~4 show that the learnable SGT works better than the handcrafted design. And promising results of two designs validate the great potential of grid representation learning paradigm.

\subsection{Grid Convolution and Its Dynamic Form}

Given a constructed grid pose $D$, now we can easily define convolution operation on grid pose, dubbed \textit{GridConv}, resembling regular convolution operations in image space.

\noindent \textbf{Vanilla form of GridConv.} Mathematically, standard GridConv operation is defined as:
\begin{equation}
	D^{out} = W*D^{in},
	\label{eqn:op}
\end{equation}
where $*$ denotes the convolution operation; $D^{in}\in \mathbb{R}^{H\times P\times C^{in}}$ and $D^{out}\in \mathbb{R}^{H\times P\times C^{out}}$ denote the input feature and the output feature, respectively; $W\in \mathbb{R}^{K\times K\times C^{in} \times C^{out}}$ denotes the convolutional kernel with kernel size $K$\texttimes$K$. With proper padding strategy, the spatial size $H$\texttimes$P$ is maintained throughout the input and output of a GridConv layer, which sharply contrasts with prevalent convolutional neural networks for image recognition tasks that typically reduce spatial feature size at multiple stages.

\begin{figure}
	\centering
	\includegraphics[width=0.9\linewidth]{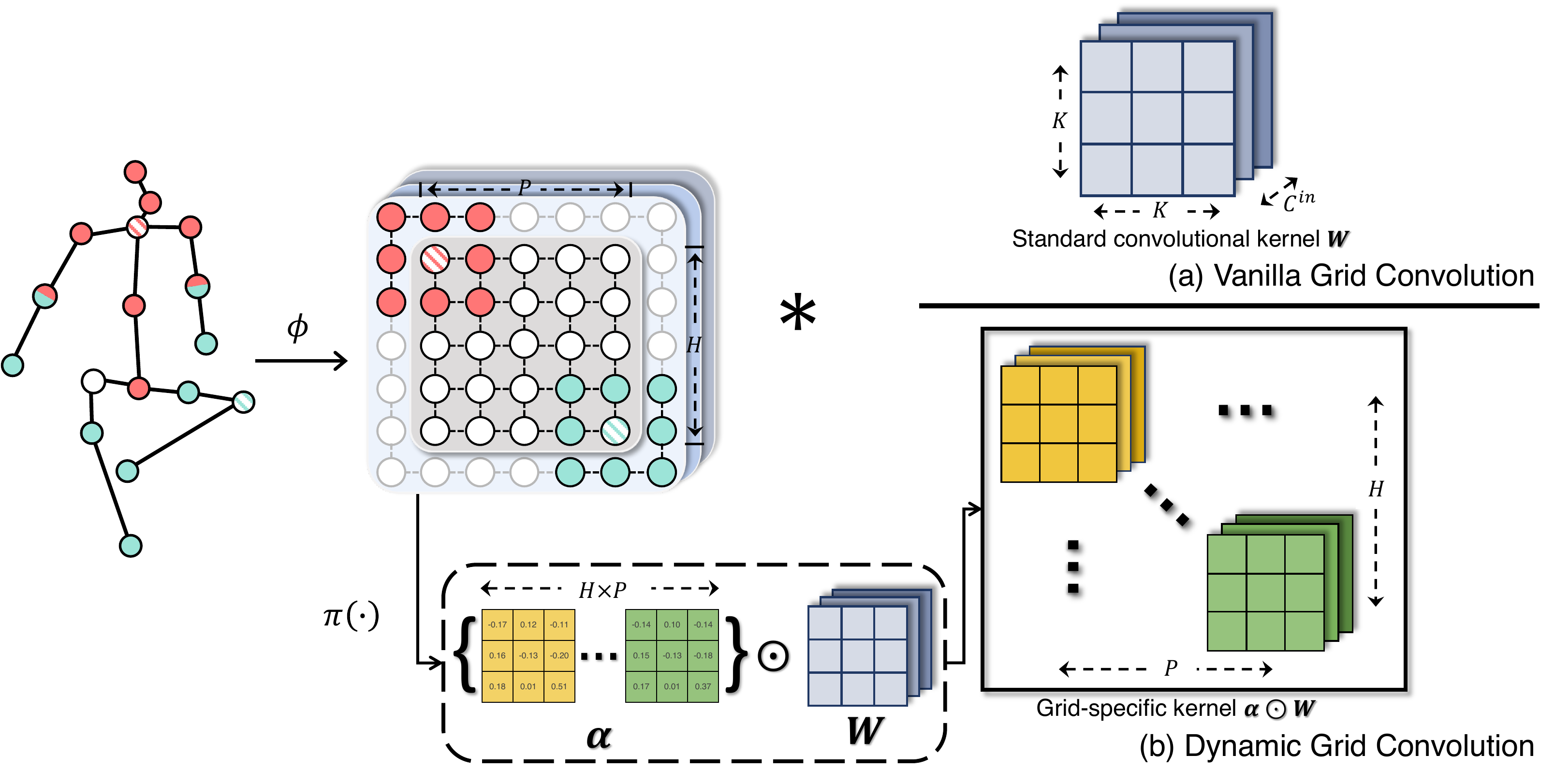}
	\caption{Illustration of (a) \textbf{vanilla GridConv} and (b) \textbf{Dynamic GridConv}. Vanilla GridConv applies the same convolutional kernel on patches, while D-GridConv makes the kernel changed according to input feature and patch location by using an attention predictor. For simplicity, we show the operation in just a single filter of the convolutional kernel.}
	\label{fig:attention}
	\vspace{-1.5em}
\end{figure}

\noindent \textbf{Dynamic form of GridConv.} According to the above definition, with vanilla GridConv, convolutional kernel $W$ is shared to the input feature, with no consideration of different grid locations or diverse body motions. To strengthen its feature learning ability on rich contextual cues, we leverage the attention mechanism conditioned on the input feature to generate attentive scaling factors to adjust the convolutional kernel, making grid convolution operations input-dependent, spatial-aware and grid-specific. We call this variant \textit{Dynamic Grid Convolution (D-GridConv)}. Specifically, following Equation~(\ref{eqn:op}), D-GridConv is defined as:
\begin{eqnarray}
	\alpha &=& \pi(D^{in}) \\
	D^{out}_{ij} &=& \left(\alpha_{ij}\odot W\right) \ast D^{in}_{\delta_{ij}},
\end{eqnarray}
where $\pi$ denotes the attention module (defined as an SE-typed structure~\cite{hu2018senet}) to generate the input-dependent scaling factor $\alpha\in \mathbb{R}^{H\times P\times K\times K}$ for adjusting the convolutional kernel $W$. Specifically, $W$ is multiplied by $\alpha_{ij}\in \mathbb{R}^{K\times K}$ on each grid patch in an element-wise manner across channel dimension. $\delta_{ij}$ denotes the index vector of local grid patch centered on grid $(i,j)$ where $i\in[1,H],j\in[1,P]$. Figure~\ref{fig:attention}a and \ref{fig:attention}b respectively illustrate how vanilla GridConv acts and how the attentive factor makes D-GridConv dynamically change with respect to grid pose.
Custom-designed attention predictor on grid convolution distinguishes D-GridConv from the series of existing attention methods, which is thoroughly discussed in our supplementary material.

\subsection{Grid Lifting Network}
With the above two components, SGT and D-GridConv, now we can construct a new category of fully convolutional lifting network in the grid pose domain, which we call \textit{Grid Lifting Network} (GLN). 
We use $\phi$ to map either detected or labeled 2D pose $G_{_{2D}}$ to 2D grid pose $D_{_{2D}}$ as the input to GLN. Then GLN uses a D-GridConv layer to expand the channel dimension of the 2D grid pose from 2 to 256, next uses two residual blocks (each incorporates two D-GridConv layers and a skip connection) to learn latent feature embedding progressively, and then uses another D-GridConv layer to shrink the channel dimension from 256 to 3, and finally gets 3D pose estimate $G_{_{3D}}$ by applying $\phi^{-1}$ over the 3D output from the last D-GridConv layer. Figure~\ref{fig:framework} shows an architectural overview of our GLN.
The entire processing pipeline of our GLN takes the following form:
\begin{equation}
	G_{_{3D}} = \phi^{-1}\left( \boldsymbol{GLN}\left(\phi\left(G_{_{2D}}\right)\right)\right).
\end{equation}

GLN is trained by minimizing the $L_2$-norm distance between the inverse 3D pose in graph structure $G_{_{3D}}$ and the ground truth pose $G^{^{GT}}_{_{3D}}$ over all training samples:
\begin{equation}
	\mathcal{L} = || G_{_{3D}} - G^{^{GT}}_{_{3D}} ||^{^2}_{_2}.
\end{equation}

Extensive experiments address the third question we raised in the Introduction section: \textit{How to construct high-performance grid lifting network (answer: GLN).}

\subsection{Relationship with GCN}
The proposed grid-structured representation learning paradigm mainly differs from GCN in two aspects:
(1) \textit{Data structure}. Grid pose encodes parent-child relation (along kinematic forward direction) and symmetry relation (along kinematic peer direction) in the bidirectional layout, yet graph pose encodes only the former one.
(2) \textit{Feature mapping scheme}. GridConv aggregates all latent channels of the neighboring nodes in a single step, yet GCN separates it into two steps recognized as latent feature mapping and neighborhood aggregation.
The single-step scheme allows GridConv to fully exploit relations of latent features. 
Please refer to our supplementary material for more implementation details and discussions.

\section{Experiments}
\label{sec:exp}

\begin{table*}\tiny
	\centering
	\resizebox{0.98\linewidth}{!}{
		\begin{tabular}{l|ccccccccccccccc|c}
			\toprule
			Method   & Dir. & Disc & Eat  & Greet & Phone & Photo & Pose & Purch. & Sit  & SitD. & Smoke & Wait & WalkD. & Walk & WalkT. & Avg. \\ \midrule
			Martinez \etal~(ICCV \citeyear{martinez2017simple}) & 51.8 & 56.2 & 58.1 & 59.0  & 69.5  & 78.4  & 55.2 & 58.1   & 74.0 & 94.6  & 62.3  & 59.1 & 65.1   & 49.5 & 52.4   & 62.9 \\
			Lee \etal~(ECCV \citeyear{lee2018propagatinglstm}) (T=1) & 43.8 & 51.7 & 48.8 & 53.1 & 52.2 & 74.9 & 52.7 & 44.6 & 56.9 & 74.3 & 56.7 & 66.4 & 47.5 & 68.4 & 45.6 & 55.8 \\
			Fang \etal~(AAAI \citeyear{fang2018posegrammar}) & 50.1 & 54.3 & 57.0 & 57.1 & 66.6 & 73.3 & 53.4 & 55.7 & 72.8 & 88.6 & 60.3 & 57.7 & 62.7 & 47.5 & 50.6 & 60.4  \\
			Zhao \etal~(CVPR \citeyear{zhao2019semgcn}) & 47.3 & 60.7 & 51.4 & 60.5  & 61.1  & \textbf{49.9}  & 47.3 & 68.1   & 86.2 & \textbf{55.0}  & 67.8  & 61.0 & \textbf{42.1}   & 60.6 & 45.3   & 57.6 \\
			Pavllo \etal~(CVPR \citeyear{pavllo2019videopose}) (T=1) & 47.1 & 50.6 & 49.0 & 51.8 & 53.6 & 61.4 & 49.4 & 47.4 & 59.3 & 67.4 & 52.4 & 49.5 & 55.3 & 39.5 & 42.7 & 51.8 \\
			Sharma \etal~(ICCV \citeyear{sharma2019ordinalranking}) & 48.6 & 54.5 & 54.2 & 55.7 & 62.6 & 72.0 & 50.5 & 54.3 & 70.0 & 78.3 & 58.1 & 55.4 & 61.4 & 45.2 & 49.7 & 58.0 \\
			Ci \etal~(ICCV \citeyear{ci2019lcn}) § & 46.8 & 52.3 & 44.7 & 50.4  & 52.9  & 68.9  & 49.6 & 46.4   & 60.2 & 78.9  & 51.2  & 50.0 & 54.8   & 40.4 & 43.3   & 52.7 \\
			Cai \etal~(ICCV \citeyear{cai2019exploitingspatialtemporal}) (T=1) & 46.5 & 48.8 & 47.6 & 50.9 & 52.9 & 61.3 & 48.3 & 45.8 & 59.2 & 64.4 & 51.2 & 48.4 & 53.5 & 39.2 & 41.2 & 50.6 \\
			Li \etal~(CVPR \citeyear{li2020cascaded})            & 47.0 & \textbf{47.1} & 49.3 & 50.5  & 53.9  & 58.5  & 48.8 & 45.5   & 55.2 & 68.6  & 50.8  & 47.5 & 53.6   & 42.3 & 45.6   & 50.9 \\
			Zeng \etal~(ECCV \citeyear{zeng2020srnet}) & 44.5 & 48.2 & 47.1 & 47.8 & 51.2 & 56.8 & 50.1 & 45.6 & 59.9 & 66.4 & 52.1 & 45.3 & 54.2 & 39.1 & 40.3 & 49.9 \\
			Zeng \etal~(ICCV \citeyear{zeng2021skeletalgcn}) § & 43.1 & 50.4 & \textbf{43.9} & 45.3 & \textbf{46.1} & 57.0 & 46.3 & 47.6 & 56.3 & 61.5 & 47.7 & 47.4 & 53.5 & \textbf{35.4} & \textbf{37.3} & 47.9 \\
			\textbf{Ours}     & 43.1 & 47.7 & 44.8 & 44.9 & 50.7 & 55.1 & 46.3 & \textbf{42.6} & 53.7 & 63.9 & 46.3 & 45.5 & 50.1 &38.6 & 40.1 & \textbf{47.6} \\
			\textbf{Ours} § & \textbf{39.9} & 47.7 & 44.7 & \textbf{43.9}  & 49.2 & 53.5 & \textbf{44.4} & 43.7 & \textbf{53.1} & 61.6 &  \textbf{45.4} & \textbf{44.7} & 47.4 & 37.7 & 37.9 & \textbf{46.3} \\
			\midrule
			Sun \etal~(ECCV \citeyear{sun2018integral}) & 47.5 & 47.7 & 49.5 & 50.2 & 51.4 & 55.8 & 43.8 & 46.4 & 58.9 & 65.7 & 49.4 & 47.8 & 49.0 & 38.9 & 43.8 & 49.6 \\
			Yang \etal~(CVPR \citeyear{yang2018adversariallearning}) & 51.5 & 58.9 & 50.4 & 57.0 & 62.1 & 65.4 & 49.8 & 52.7 & 69.2 & 85.2 & 57.4 & 58.4 & 43.6 & 60.1 & 47.7 & 58.6 \\
			Moon \etal~(ICCV \citeyear{moon2019camdisaware}) & 50.5 & 55.7 & 50.1 & 51.7 & 53.9 & 55.9 & 46.8 & 50.0 & 61.9 & 68.0 & 52.5 & 49.9 & 41.8 & 56.1 & 46.9 & 53.3 \\
			Moon \etal~(ECCV \citeyear{moon2020meshnet}) & - & - & - & - & - & - & - & - & - & - & - & - & - & - & - & 55.7 \\
			Zhou \etal~(PAMI \citeyear{zhou2021hemlets}) & \textbf{34.4} & \textbf{42.4} & \textbf{36.6} & \textbf{42.1} & \textbf{38.2} & \textbf{39.8} & \textbf{34.7} & \textbf{40.2} & \textbf{45.6} & 60.8 & \textbf{39.0} & \textbf{42.6} & \textbf{42.0} & \textbf{29.8} & \textbf{31.7} & \textbf{39.9} \\ 
			\textbf{Ours} - joint training & 36.9  & 44.4 & 41.9 & 43.3 & 45.6  & 47.8  & 43.0 & 40.7 & 50.7 & \textbf{60.6} & 44.3   & 43.6  & 43.9 & 33.9 & 35.0  & 43.7 \\ \bottomrule
		\end{tabular}
	}
	\caption{MPJPE comparison (mm) of our method against both mainstream lifting and end-to-end methods on Human3.6M. For the comparison with lifting methods (upper half of the table), we report the results under Protocol 1 using 2D detection input. T=1 denotes single-frame results of temporal methods. § denotes estimating 2D pixel and 3D depth jointly. For the comparison with end-to-end methods (lower half of the table), we report the results under image input.}
	\label{tab:accuracy_action}
	\vspace{-1.2em}
\end{table*}

\begin{table}[t]\tiny
	\centering
	\begin{tabular}{l|c|ccc}
		\toprule
		\multicolumn{1}{c|}{\multirow{2}{*}{Method}} & \multicolumn{1}{c|}{\multirow{2}{*}{Special Mark}} & \multicolumn{3}{c}{MPJPE} \\
		\multicolumn{1}{c|}{}                 &       & P1      & P1*    & P2     \\ \midrule
		Martinez \etal~(ICCV \citeyear{martinez2017simple})
		& -   & 62.9    & 45.5   & 47.7   \\
		Zhao \etal~(CVPR \citeyear{zhao2019semgcn})          & -       & 57.6    & 43.8   & -   \\
		Fang \etal~(AAAI \citeyear{fang2018posegrammar}) & - & 60.4 & - & 45.7 \\
		Sharma \etal~(ICCV \citeyear{sharma2019ordinalranking}) & -   & 58.0    & -      & 40.9   \\
		Pavllo \etal~(CVPR \citeyear{pavllo2019videopose}) & T=1 & 51.8 & - & 40.0 \\
		Ci \etal~(ICCV \citeyear{ci2019lcn})             & §          & 52.7   & 36.3  & 42.2   \\
		Cai \etal~(ICCV \citeyear{cai2019exploitingspatialtemporal}) & T=1 & 50.6 & 38.1 & 40.2 \\
		Zeng \etal~(ECCV \citeyear{zeng2020srnet})      & -        & 49.9    & 36.4   & -      \\
		Li \etal~(CVPR \citeyear{li2020cascaded}) & -      & 50.9    & 34.5   & 38.0   \\
		Yu \etal~(CVPR \citeyear{yu2021pcls})           &  -       & 67.0    & 40.1   & -      \\
		Gong \etal~(CVPR \citeyear{gong2021poseaug})	& 16 joints & 50.2 & 36.9 & 39.1 \\
		Zeng \etal~(ICCV \citeyear{zeng2021skeletalgcn}) & § & 47.9    & 30.4   & 39.0   \\
		\midrule
		\textbf{Ours}                                      & -         & 47.6 & 36.4 & \textbf{37.4}         \\
		\textbf{Ours}                                     & §              & \textbf{46.3} & \textbf{29.5} & 37.6\\
		\bottomrule
	\end{tabular}
	\caption{Comparison on Human3.6M under all protocols. § denotes estimating 2D pixel and 3D depth jointly.}
	\label{tab:accuracy_protocol}
	\vspace{-2.8em}
\end{table}

\begin{table}[t]\tiny
	\centering
	\begin{tabular}{l|c|ccc}
		\toprule
		Method       & Cross evaluation & PCK\textuparrow           & AUC\textuparrow           & MPJPE\textdownarrow         \\ \midrule
		Mehta \etal~\citeyear{mehta2017improvedcnnsup_mpiinf3dhp}       & \xmark   & 76.5          & 40.8          & 117.6         \\
		Mehta \etal~\citeyear{mehta2017vnect}       & \xmark   & 76.6          & 40.4          & 124.7         \\
		LCR-Net~\citeyear{rogez2017lcrnet}     & \xmark   & 59.6          & 27.6          & 158.4         \\
		Zhou \etal~\citeyear{zhou2017towards}         & \xmark   & 69.2          & 32.5          & 137.1         \\
		Multi Person~\citeyear{mehta2018multiperson} & \xmark   & 75.2          & 37.8          & 122.2         \\
		OriNet~\citeyear{luo2018orinet}       & \xmark   & 81.8          & 45.2          & \;\,89.4 \\
		\midrule
		HMR~\citeyear{kanazawa2018hmr}     &  \cmark  & 77.1          & 40.7          & 113.2         \\
		Yang \etal~\citeyear{yang2018adversariallearning}         &  \cmark  & 69.0          & 32.0          & -             \\
		Li \etal~\citeyear{li2019multihypotheses}           &  \cmark  & 67.9          & -             & -             \\
		LCN~\citeyear{ci2019lcn} & \cmark & 74.0 & 36.7 & - \\
		Li \etal~\citeyear{li2020cascaded}           &  \cmark  & 81.2          & 46.1          & \;\,99.7          \\
		SRNet~\citeyear{zeng2020srnet} & \cmark & 77.6 & 43.8 & - \\
		SkeletalGCN~\citeyear{zeng2021skeletalgcn}   & \cmark & 82.1 & 46.2 & - \\
		PoseAug~\citeyear{gong2021poseaug} & \cmark &  88.6 & 57.3 & \;\,73.0 \\
		\textbf{Ours}        &  \cmark  & \textbf{89.2} & \textbf{57.6} & \;\,\textbf{72.1} \\
		\bottomrule
	\end{tabular}
	\caption{Performance comparison on MPI-INF-3DHP.}
	\label{tab:3dhp}
\end{table}

\subsection{Datasets}
\noindent \textbf{Human3.6M}. It is the largest indoor 3D human motion benchmark with 3D labels collected by motion capture system~\cite{h36m_pami}.
The dataset consists of 11 actors playing a variety of activities. 
We follow the convention that takes \textit{Subject 1,5,6,7,8} as the training set and \textit{Subject 9,11} as the evaluation set.
We measure the result by Mean Per Joint Position Error (MPJPE) in millimeters under three protocols.
\textbf{Protocol 1 (P1)} takes 2D pose detection from HRNet~\cite{sun2019hrnet} as input.
\textbf{Protocol 1* (P1*)} takes ground truth 2D pose as input.
\textbf{Protocol 2 (P2)} takes 2D pose detection as input and measures 3D error after aligning 3D estimate to the ground truth through rigid alignment.

\noindent \textbf{MPI-INF-3DHP}. It is another 3D human motion benchmark with 3D labels obtained by multi-view reconstruction~\cite{mehta2017improvedcnnsup_mpiinf3dhp}.
To evaluate the generalization ability of our method, we consider challenging cross-dataset evaluation, applying the model trained on Human3.6M for direct test on the evaluation set of MPI-INF-3DHP.
The evaluation metrics include MPJPE, Percentage of Correct Keypoints (PCK), and Area Under the Curve (AUC) of PCK.

\subsection{Implementation Details}
Considering that the number of body joints $J$ popularly used for these two datasets is 17, the size of grid pose $H$\texttimes$P$ should be no smaller than 6\texttimes3 or 5\texttimes4 due to $HP$$\geq$$J$. 
We use a grid pose with 5\texttimes 5 size as our default setting.

In the grid lifting network, each D-GridConv layer is composed of two-branch D-GridConv, batch normalization, ReLU, and dropout operations. The attention module of D-GridConv consists of global average pooling followed by batch normalization and ReLU, two linear layers (reducing channel dimension first to 16 and further to 3), and a Sigmoid activation function.
The convolutional kernel size is fixed to 3\texttimes3.
Two-branch D-GridConv divides the feature extraction into two branches, applying grid convolution on circular padded and replicate padded grid pose respectively, and finally outputs the sum of their results.

We train the model with Adam optimizer using a batch size of 200 and a learning rate starting at $0.001$ for 100 epochs. In AutoGrids, $S^{prob}$ is initialized by handcrafted SGT for 5\texttimes5 grid and by random value for other sizes. 
We stop adding Gumbel noise at the 30th epoch to slow down the rate of grid pose changes. Our model has totally 4.79 million learnable parameters with 0.04 million from the attention modules and \textless 1k learnable parameters from AutoGrids. We train and test the model on a single NVIDIA 1080Ti GPU. Commonly, one run of model training takes about 40 hours, and the runtime speed of our model is over 1600 FPS.

\label{sec:performance}

\subsection{Comparison with State-of-The-Art Methods}
First, we describe experimental comparisons under conventional single-dataset evaluation on Human3.6M and challenging cross-dataset evaluation on MPI-INF-3DHP.

\noindent \textbf{Results on Human3.6M}.
In the upper half part of Table~\ref{tab:accuracy_action}, we compare our method with mainstream lifting methods on Human3.6M. Note that two works~\cite{ci2019lcn,zeng2021skeletalgcn} marked by § first predict 2D pixel coordinate and a depth value for each joint, and then post-process them into 3D physical coordinate with given camera intrinsics and body root position in camera space. For a fair comparison with them, we also report our result §, showing remarkable gains (\textgreater1.6\,$mm$). For those approaches dealing with temporal 2D pose input, we report their single-frame results. 
We can see that our method achieves the best MPJPE of 47.6\,$mm$ compared to existing lifting works. And data normalization strategy § pushes our performance further to 46.3\,$mm$, showing great margins against lifting methods.

In the lower half part of Table~\ref{tab:accuracy_action}, we jointly train the 2D detector and our grid lifting network in an end-to-end manner, and compare our method to mainstream end-to-end methods.
Joint training improves our method to correct erroneous 2D detection results and shows significant model performance improvements against the lifting-alone training. Besides, our model from joint training outperforms most end-to-end methods and approaches state of the art.

We further evaluate our method under all three protocols, and summarize the performance comparison in Table~\ref{tab:accuracy_protocol}. Generally, our method outperforms most of existing works under 2D ground truth input (P1*). When adopting data normalization strategy §, our method is superior to state of the arts under both 2D detection input and GT input.

\noindent \textbf{Results on 3DHP}. To better explore the generalization ability of our method, we compare it with previous works that adopt cross-dataset evaluation.
Detailed results are shown in Table~\ref{tab:3dhp}, in which we also include existing works  performing both from-scratch training and evaluation on 3DHP,
in order to have a more comprehensive comparison.
We can observe that our method outperforms all methods by significant margins with respect to all three metrics.

\subsection{Qualitative Results}
\label{sec:qualitative}
\begin{figure}
	\centering
	\begin{subfigure}{\linewidth}
		\centering
		\includegraphics[width=0.95\linewidth]{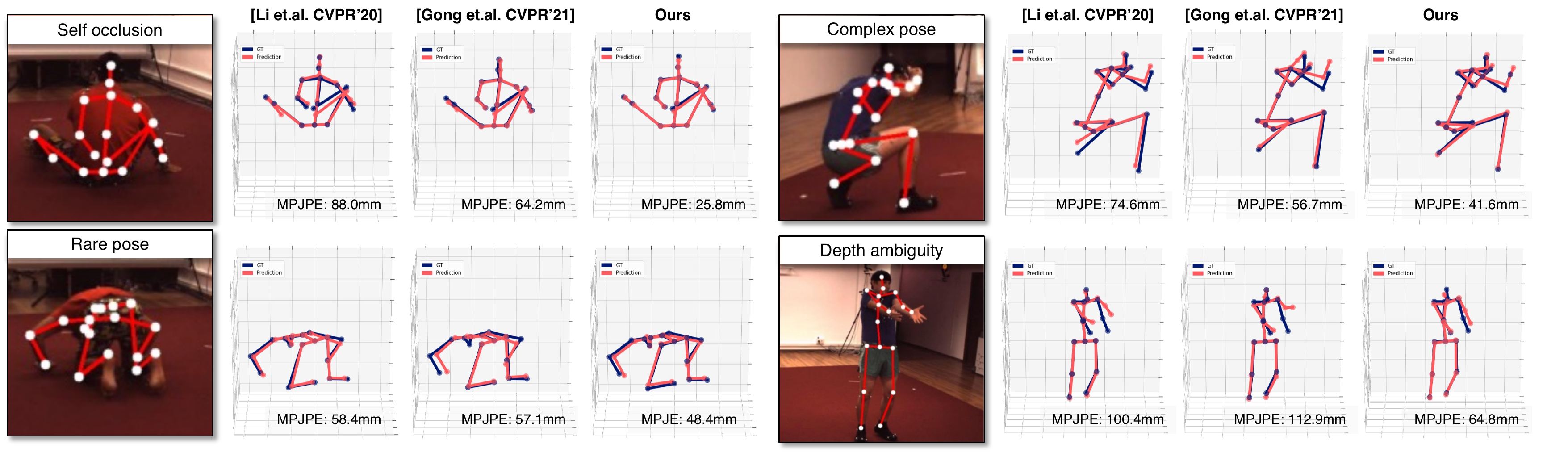}
		\caption{Evaluation results on Human3.6M test set.}
		\label{fig:vis_hm36}
	\end{subfigure}
	\begin{subfigure}{\linewidth}
		\centering
		\includegraphics[width=0.95\linewidth]{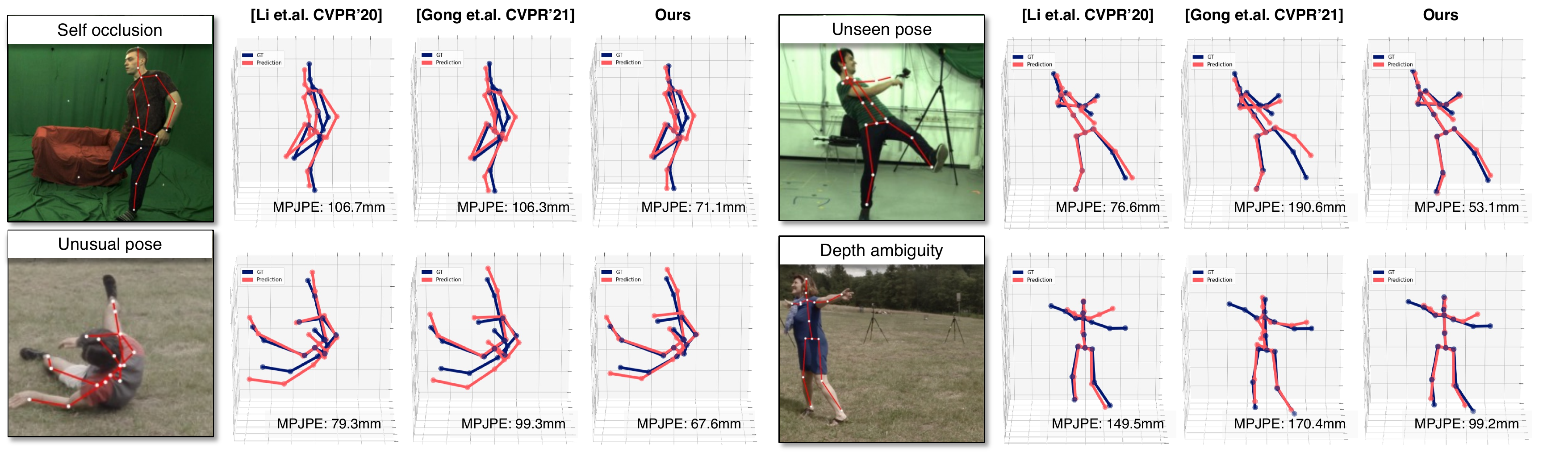}
		\caption{Cross-dataset results on MPI-INF-3DHP test set.}
		\label{fig:vis_3dhp}
	\end{subfigure}
	\vspace{-0.5em}
	\caption{Visualization comparison between top-performing methods and GLN.}
	\label{fig:vis}
	\vspace{-1em}
\end{figure}

Next, we provide some qualitative comparisons of our best-performed model trained on Human3.6M to illustrate the ability to handle challenging scenarios with various viewpoints and severe occlusions. Figure~\ref{fig:vis_hm36} and \ref{fig:vis_3dhp} show visualization results respectively on Human3.6M and on 3DHP. 

\subsection{Ablation Study} \label{sec:ablation}
Finally, we provide a lot of ablative experiments to study different components and design aspects of our GLN. All ablative experiments are performed on Human3.6M dataset. More experimental setups, ablations, and discussions can be found in our supplementary material.

\begin{table}\small
	\centering
	\resizebox{0.8\linewidth}{!}{
		\begin{tabular}{p{0.36\linewidth} | >{\centering}p{0.14\linewidth}>{\centering}p{0.14\linewidth}>{\centering\arraybackslash}p{0.14\linewidth}}
			\toprule
			Method & Original & $G_{craft} $ & $D_{craft}$ \\ \midrule
			SemGCN~\shortcite{zhao2019semgcn} & 43.8 & \underline{41.8} & 43.5 \\
			LCN~\shortcite{ci2019lcn} & 39.7 & \underline{39.4} & 39.5 \\
			GridConv & - & - & 39.0 \\
			D-GridConv & - & - & \textbf{37.1} \\ \bottomrule
		\end{tabular}
	}
	\caption{Ablation study of applying grid pose on GCN methods. We report MPJPE under ground truth input. $G_{craft}$ denotes graph pose merging grid topology and $D_{craft}$ denotes handcrafted grid pose. Underline marks the best of the row.}
	\label{tab:adj_mat}
	\vspace{-1.5em}
\end{table}

\noindent \textbf{Grid versus Graph.}
Being composed of nodes and edges, grid pose has a similar structure to graph pose, which makes it possible to be combined in graph convolution framework.
It challenges the significance of proposing grid convolution.
So we investigate the combination by setting two kinds of input pose:
(1) graph-structured pose having handcrafted-grid topology, denoted as $G_{craft}$; (2) handcrafted grid pose, denoted as $D_{craft}$.
We select two GCN baselines LCN~\shortcite{ci2019lcn} and SemGCN~\shortcite{zhao2019semgcn}. For the experiment using $D_{craft}$, we modify their convolution layers to receive 5\texttimes5 grid input. 
Results shown in Table~\ref{tab:adj_mat} demonstrate three facts.
First, results on $G_{craft}$ (43.8\textrightarrow41.8, 39.7\textrightarrow39.4) indicate that grid-based topology is helpful to pose learning. 
Second, results on $D_{craft}$ indicate that employing grid pose on GCN does not ensure better performance (41.8\textrightarrow43.5, 39.4\textrightarrow39.5).
Last, GridConv family performing better indicates that grid convolution is more effective and more suitable for grid pose than lifting-based graph convolution operators.

\noindent \textbf{SGT designs for constructing grid pose.}
Although we provide both handcrafted and learnable SGT designs, a more straightforward SGT design is to generate a grid pose randomly. Hence it is necessary to compare their effectiveness.
Accordingly, we conducted a set of ablative experiments using these three designs separately to construct 5\texttimes5 grid pose, and report results in Table~\ref{tab:rand_org}. When generating a random grid pose, each joint is forced to appear at least once. Surprisingly, it can be observed that random grid layouts show good performance even with no semantic skeleton topology constraint contained. Comparatively, our handcrafted and learnable designs are obviously better than random ones. These results echo \textit{merits of SGT} of Section~3.1.

\begin{table}\tiny
	\centering
	\begin{tabular}{l|lll}
		\toprule
		Grid Pose by & \quad\,\,P1 & \,\,P1* & \;P2 \\ \midrule
		Random SGT \#1                    & \quad\underline{49.3}    & 38.5   & 38.5   \\
		Random SGT \#2                    & \quad49.5    & 38.2   & 38.5   \\
		Random SGT \#3                    & \quad49.5    & 37.9   & \underline{38.3}   \\
		Random SGT \#4                    & \quad49.6    & \underline{37.8}   & 38.4   \\
		\textit{Mean over} \#1-4                 & \quad49.5    & 38.1   & 38.4   \\ \midrule
		Handcrafted SGT                 &\quad47.9$_{_{2.8\%\downarrow}}$    &37.1$_{_{1.9\%\downarrow}}$   &37.9$_{_{1.0\%\downarrow}}$   \\
		Learnable SGT                          & \quad47.6$_{_{0.6\%\downarrow}}$        & 36.4$_{_{1.9\%\downarrow}}$   &  37.4$_{_{1.3\%\downarrow}}$      \\\bottomrule
	\end{tabular}
	\caption{Ablation study on GLN using different SGT designs to construct grid pose. The size of grid pose is fixed to 5\texttimes 5.}
	\label{tab:rand_org}
	\vspace{-2em}
\end{table}

\begin{figure}[t]
	\centering
	\begin{subfigure}{0.26\linewidth}
		\centering
		\includegraphics[height=1.1\linewidth]{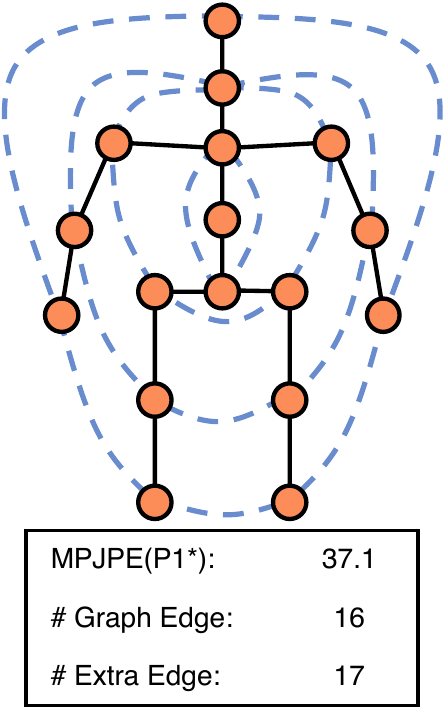}
		\caption{Handcrafted.}
	\end{subfigure}
	\begin{subfigure}{0.26\linewidth}
		\centering
		\includegraphics[height=1.1\linewidth]{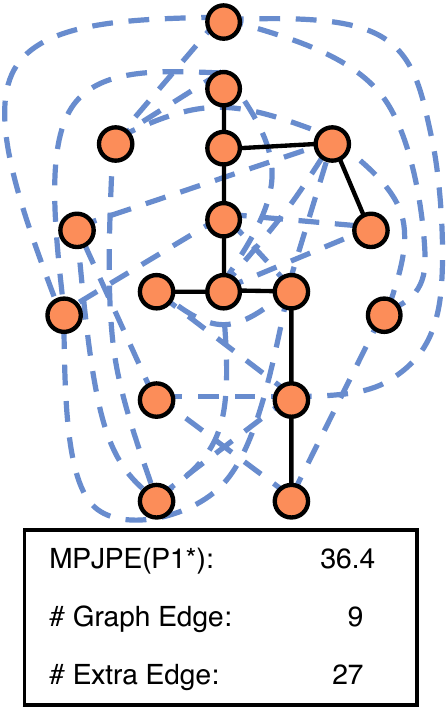}
		\caption{Learnt \#1.}
	\end{subfigure}
	\begin{subfigure}{0.26\linewidth}
		\centering
		\includegraphics[height=1.1\linewidth]{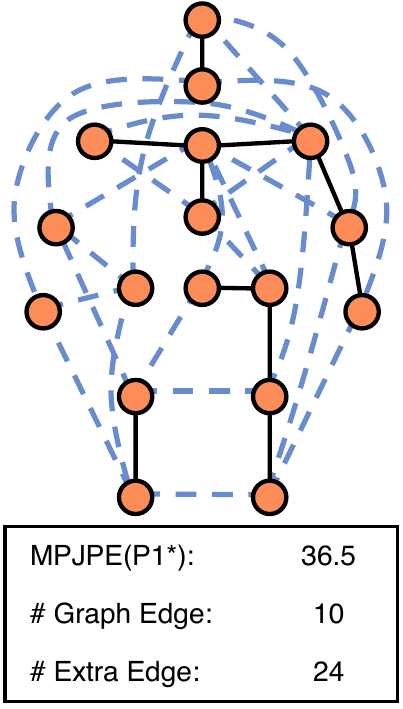}
		\caption{Learnt \#2.}
	\end{subfigure}
	\vspace{-0.5em}
	\caption{Visualization of handcrafted and learnt grid pose patterns converted into an equivalent graph structure. The connected joints are neighboring in the grid pose.}
	\label{fig:skeleton_vis}
	\vspace{-0.5em}
\end{figure}

\begin{table}[t]\small
	\centering
	\resizebox{0.65\linewidth}{!}{
		\begin{tabular}{l|ccc}
			\toprule
			\multicolumn{1}{l|}{\begin{tabular}[l]{@{}l@{}}Body Joint \# (J)\\ Grid Pose Size ($H$\texttimes$P$)\end{tabular}} & \begin{tabular}[l]{@{}c@{}}17\\ 5\texttimes 5\end{tabular} & \begin{tabular}[l]{@{}c@{}}32\\ 7\texttimes 5\end{tabular} & $\Delta$ \\ \midrule
			Handcrafted SGT  & 37.1 & \textbf{35.0}  & 2.1 \\
			Learnable SGT    & 36.4   & \textbf{33.4}  & 3.0   \\ \bottomrule
		\end{tabular}
	}
	\caption{Ablation study on using different numbers of body joints as input. We report MPJPE under ground truth input.}
	\label{tab:32j}
	\vspace{-1em}
\end{table}

\noindent \textbf{Analysis of learnt grid pose patterns.} So far, we can conclude that the grid pose by learnable SGT design performs better than both handcrafted and random ones. 
To have a better understanding of learnable SGT design, in Figure~\ref{fig:skeleton_vis}, we visualize two learnt grid pose patterns converted into an equivalent graph structure where dotted edges denote neighboring joints in grid pose. We can notice that the learnt grid pose patterns maintain fewer skeleton edges, yet establish new edges to keep all graph nodes reachable, which include many long-distance connections (e.g. head to knees). 

\begin{figure}[H]
	\begin{minipage}{0.48\linewidth}
		\centering
		\includegraphics[height=0.65\linewidth]{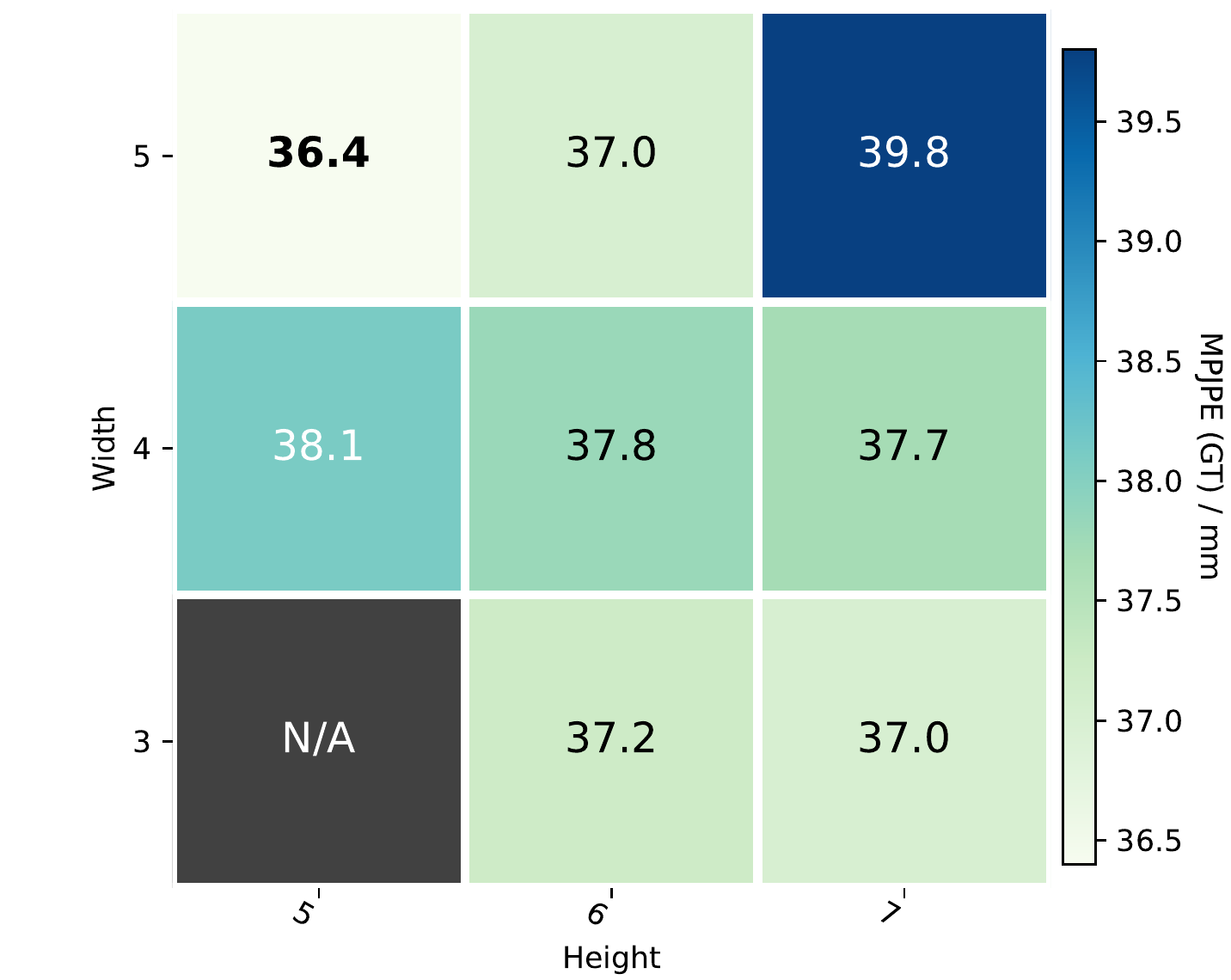}
		\caption{Ablation study on the effect of grid pose with different $H$\texttimes$P$ settings.}
		\label{fig:grid_size}
	\end{minipage}
	\hfill
	\begin{minipage}{0.48\linewidth}
		\centering
		\includegraphics[height=0.65\linewidth]{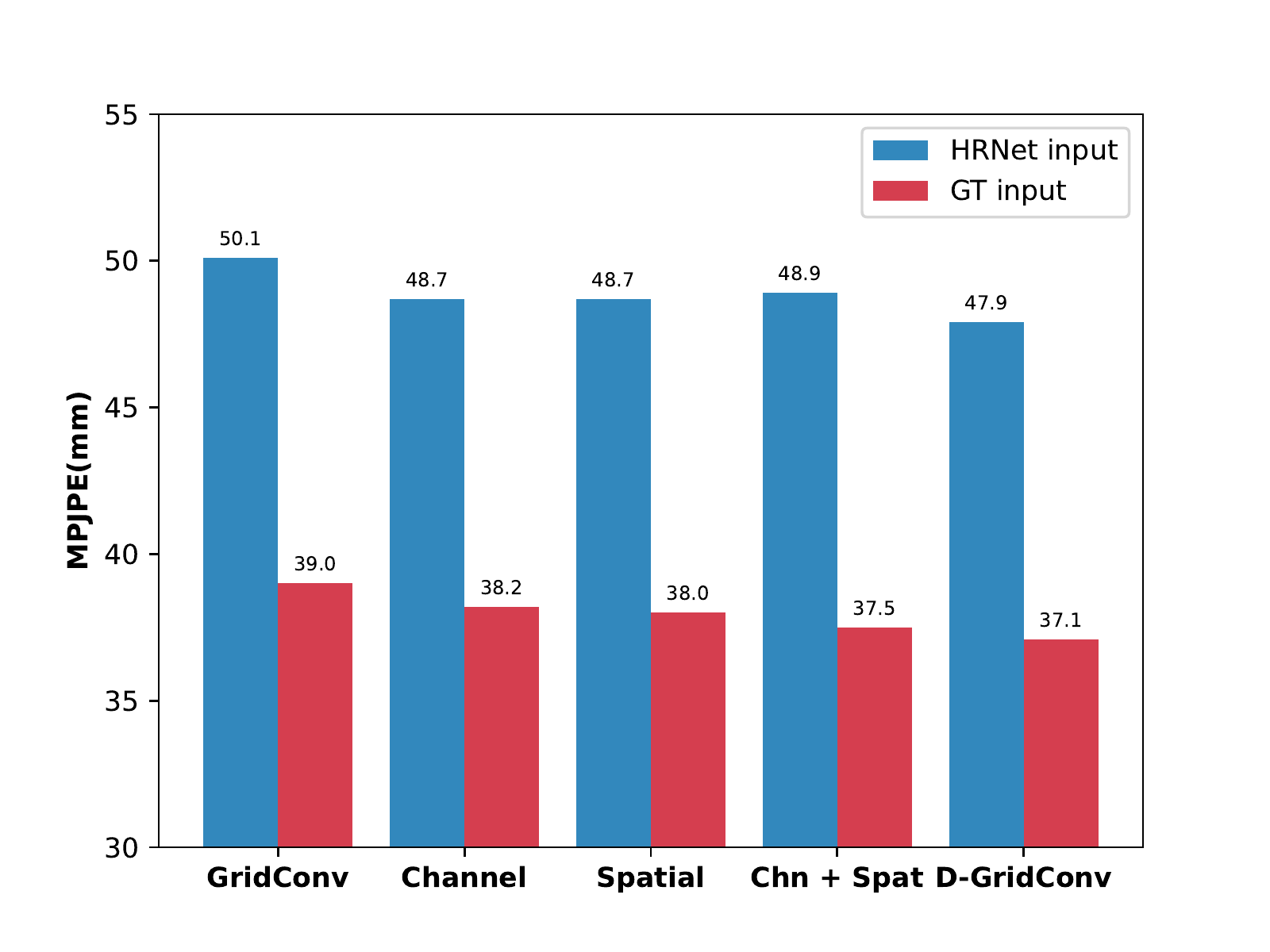}
		\caption{Ablation study on different attention designs for grid convolution.}
		\label{fig:atten_comp}
	\end{minipage} 
\end{figure}

\noindent \textbf{Grid size.}
When using SGT to construct a grid pose, a critical question is how to select a proper size $H$\texttimes$P$ for grid pose. Accordingly,
we conducted an experiment to compare the performance of training our lifting network with different $H$\texttimes$P$ settings of grid pose.
Detailed results are shown in Figure~\ref{fig:grid_size}. It demonstrates that 5\texttimes 5 size reaches the best performance, hence is set as the default.

\noindent \textbf{Grid pose with different body joint numbers.} 
Note that a skeleton pose with 17 body joints is used in our main experiments, following many existing works. 
We also performed an experiment to investigate the generalization ability of our method with another skeleton pose having 32 body joints. 
The additional joints include hands and feet that enrich motion information and make the task more challenging.
As shown in Table~\ref{tab:32j}, the results demonstrate that additional joint information helps our model to learn a more accurate 3D pose, which indicates our method has the potential to handle more complicated skeleton pose patterns.

\noindent \textbf{Attention designs.}
Regarding the attention module for D-GridConv, we also tried several other designs besides the proposed one including:
(a) channel-wise attention, similar to SENet~\cite{hu2018senet};
(b) spatial-wise attention, a reduced version of CBAM~\cite{woo2018cbam};
(c) spatial+channel attention, similar to CBAM.
Figure~\ref{fig:atten_comp} compares the performance, showing our design performs the best.

\section{Conclusion}
In this paper, we take the lead in extending convolution operations to estimate 3D human pose from 2D pose detection by shifting pose representation from graph structure to weave-like grid pose through a novel transformation called Semantic Grid Transformation (SGT).
We offer two paths to implement SGT, namely handcraft SGT and learnable SGT.
Based on the grid layout, we formulate grid convolution and construct a grid lifting network with its attentive variant.
Extensive experiments on two public benchmarks demonstrate the superiority of our method to state-of-the-art works.

\section*{Acknowledgements}
The authors would like to acknowledge the support from NSFC (No. 62072449, No. 61972271) and Macau Sci and Tech Fund (0018/2019/AKP).

\bibliographystyle{aaai23}
\bibliography{reference}

\newpage
\appendix



\appendix
\renewcommand{\thetable}{\Alph{table}}
\renewcommand{\thefigure}{\Alph{figure}}
\setcounter{figure}{0}
\setcounter{table}{0}
\begin{table*}[ht]\tiny
	\centering
	\resizebox{0.98\linewidth}{!}{%
		\begin{tabular}{l|ccccccccccccccc|c}
			\toprule
			Method   & Dir. & Disc & Eat  & Greet & Phone & Photo & Pose & Purch. & Sit  & SitD. & Smoke & Wait & WalkD. & Walk & WalkT. & Avg. \\ \midrule
			Martinez \etal~(ICCV'17)~\cite{martinez2017simple} & 39.5 & 43.2 & 46.4 & 47.0  & 51.0  & 56.0  & 41.4 & 40.6   & 56.5 & 69.4  & 49.2  & 45.0 & 49.5   & 38.0 & 43.1   & 47.7 \\
			Fang \etal~(AAAI'18)~\cite{fang2018posegrammar} & 38.2 & 41.7 & 43.7 & 44.9 & 48.5 & 55.3 & 40.2 & 38.2 & 54.5 & 64.4 & 47.2 & 44.3 & 47.3 & 36.7 & 41.7 & 45.7  \\
			Pavllo \etal~(CVPR'19)~\cite{pavllo2019videopose}~(T=1) & 36.0 & 38.7 & 38.0 & 41.7 & 40.1 & 45.9 & 37.1 & 35.4 & 46.8 & 53.4 & 41.4 & 36.9 & 43.1 & 30.3 & 34.8 & 40.0 \\
			Sharma \etal~(ICCV'19)~\cite{sharma2019ordinalranking} & 35.3 & 35.9 & 45.8 & 42.0 & 40.9 & 52.6 & 36.9 & 35.8 & 43.5 & 51.9 & 44.3 & 38.8 & 45.5 & \textbf{29.4} & 34.3 & 40.9 \\
			Ci \etal~(ICCV'19)~\cite{ci2019lcn} § & 36.9 & 41.6 & 38.0 & 41.0 & 41.9 & 51.1 & 38.2 & 37.6 & 49.1 & 62.1 & 43.1 & 39.9 & 43.5 & 32.2 & 37.0 & 42.2\\
			Cai \etal~(ICCV'19)~\cite{cai2019exploitingspatialtemporal}~(T=1) & 36.8 & 38.7 & 38.2 & 41.7 & 40.7 & 46.8 & 37.9 & 35.6 & 47.6 & \textbf{51.7} & 41.3 & 36.8 & 42.7 & 31.0 & 34.7 & 40.2 \\
			Zeng \etal~(ICCV'21)~\cite{zeng2021skeletalgcn} § & 33.9 & 37.2 & 36.8 & 38.1 & \textbf{38.7} & \textbf{43.5} & 37.8 & 35.0 & 47.2 & 53.8 & 40.7 & 38.3 & 41.8 & 30.1 & \textbf{31.4} & 39.0 \\
			\midrule
			\textbf{Ours}  & \textbf{33.1} & \textbf{37.2} & \textbf{35.7} & 36.3 & 39.3 & 43.8 & \textbf{34.4} & \textbf{33.1} & \textbf{43.0} & 52.5 & \textbf{37.4} & 34.8 & 39.4 & 29.5 & 32.2 & \textbf{37.4} \\
			\textbf{Ours} § & 33.3 & 37.9 & 36.4 & \textbf{36.2} & 39.0 & 44.6 & \textbf{34.4} & 34.3 & 43.2 & 52.5 & \textbf{37.4} & \textbf{34.7} & \textbf{38.9} & 29.5 & 32.0 & 37.6\\ 
			\bottomrule
	\end{tabular}}
	\caption{Performance comparison regarding PA-MPJPE with rigid alignment from the ground truth under Protocol 2 on Human3.6M. T=1 denotes single-frame results of temporal methods. § denotes predicting 2D pixel coordinate and 3D depth.}
	\label{tab:pa-mpjpe}
\end{table*}

\section{More Details about Implementation}			

\subsection{Inverse SGT Process}
In the main paper, we did not describe inverse SGT process in detail due to limited space.
Here we present its formulation as follows:
\begin{eqnarray}
	\label{eqn:avg}
	\tilde{S}^T_i  &=& \frac{S^T_i}{\sum_{j=1}^{HP} S^T_{ij}} \\
	\label{eqn:inv_sgt}
	\phi^{-1}(D) &=& \tilde{S}^T \times D
\end{eqnarray}
Considering $HP$$\geq$$J$, a $H$\texttimes$P$ grid pose may have multiple proposal values for a certain joint.
For inverse SGT process, directly adding these proposals up as the body joint feature would produce magnitude imbalance among joints.
So we propose to take the mean value of proposals as the joint feature by normalizing the transpose of assignment matrix $S^T$ with Equation \ref{eqn:avg}.

\subsection{Network Construction} Each D-GridConv layer consists of two-branched convolution operators.
In both of the branches, $H$\texttimes $P$ input grid pose is first padded to $H_{pad}$\texttimes $P_{pad}$ with padding size of $\frac{K-1}{2}$ on each side, where $K$ is the kernel size.
The padding content for the first branch is circular value and that for the second branch is replicate value.
After padding, convolution operators take the padded grid pose as input and produce $H$\texttimes$P$ output.
Finally, the outputs of the two branches are added up as the output of the current D-GridConv layer.


\subsection{Training Details}
Dropout probability is set to 0.25.
When training with a handcrafted grid pose, the learning rate starts at 0.001 and decays per epoch by a scaling factor of 0.96.
When training with AutoGrids, the learning rate still starts at 0.001 but decays per 10 epochs by a scaling factor of 0.1.
During the first 30 epochs of training, we enable Gumbel noise in AutoGrids to encourage exploration of grid pose patterns. The temperature in Gumbel Softmax function is fixed to 1.
During this period, probability values in $S^{prob}$ change dramatically and the grid pose pattern is not determined.
After 30 epochs, Gumbel noise is disabled, but the probability values are still updated with a small magnitude.
It is worth noting that AutoGrids does not guarantee that the binary assignment matrix $S$ maps all body joints into the grid pose during the learning process.
However, the loss indicator would guide the optimizing direction for probability matrix $S^{prob}$ to generate a fully projected $S$.
As a result, AutoGrids keeps searching for a locally optimal grid pattern and finally locates the best one.
Notably, the best grid pose pattern is not unique, as shown in the learnt grid pose patterns \#1 and \#2 of Figure 7 in the main paper.

\subsection{Data Normalization} 
2D and 3D pose data are normalized by constant scaling factors. On Human3.6M and MPI-INF-3DHP datasets, 2D pixel coordinate is scaled to $[-1, 1]$, and 3D physical coordinate in millimeter is divided by 1000.
When using another data normalization strategy (marked as §) that predicts 2D pixel coordinate $(u,v)$ and 3D depth $z$, we directly follow the convention proposed in \cite{ci2019lcn}. 
During training, the supervision is conducted on $(u,v,z)$.
During the evaluation, estimated $(u,v)$ in 2D is projected to $(x,y)$ in 3D through the perspective camera model using known camera focal lens, optical center, and 3D offset coordinate from body root to the camera.
Then MPJPE is measured on post-processed $(x,y,z)$.

\begin{figure*}
	\centering
	\begin{subfigure}{0.3\linewidth}
		\centering
		\includegraphics[height=8.0em]{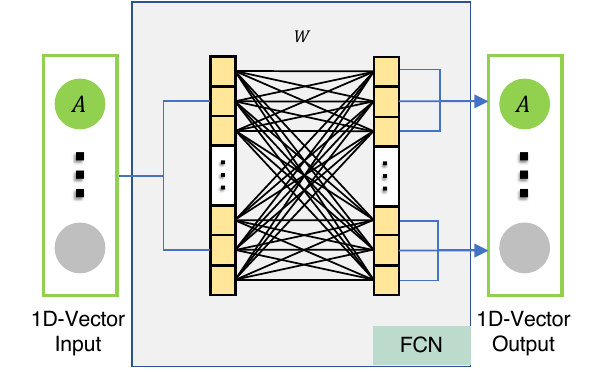}
		\caption{Fully-connected feature mapping of linear operator.}
	\end{subfigure}
	\quad
	\begin{subfigure}{0.3\linewidth}
		\centering
		\includegraphics[height=8.0em]{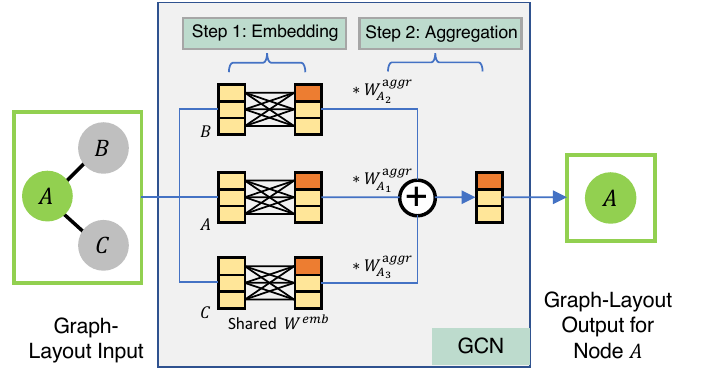}
		\caption{Two-step feature spatial aggregation and embedding of graph convolution operator.}
	\end{subfigure}
	\quad
	\begin{subfigure}{0.3\linewidth}
		\centering
		\includegraphics[height=8.0em]{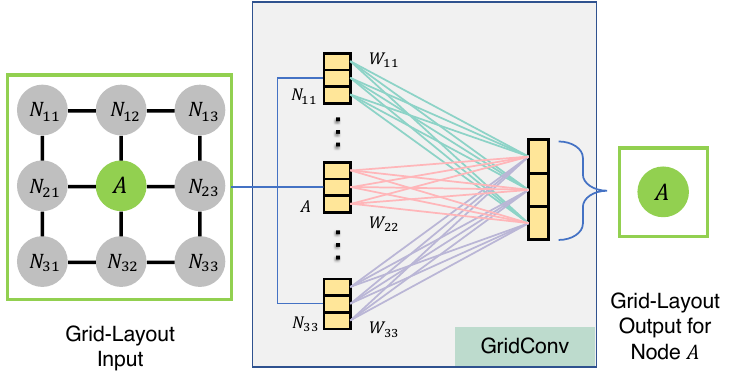}
		\caption{Single-step feature mapping of grid convolution operator.}
	\end{subfigure}
	\caption{Conceptual difference of FCN, GCN and GridConv.}
	\label{fig:fcn_gcn_gridconv}
\end{figure*}

\subsection{Relationship with FCN and GCN}
In the main paper, we briefly discuss the relationship between GridConv and GCN.
Here we first provide comprehensive formulation comparisons on FCN, GCN, and GridConv.
The differences mainly exist in two aspects: \textit{data structure of human pose} and \textit{feature mapping scheme of the neural operator}.
Below we discuss them in turn.

\noindent\textbf{Data structure}. 1D vector representation, the standard feature representation for FCN, is obtained by discarding skeleton topology and flattening the pose feature.
Being considered as the intuitive skeleton representation for human pose, graph structure usually encodes joint 3D coordinates in nodes and parent-child relation in edges.
Yet motion symmetry is not explicitly taken into account due to no edge connections between peer joints.
The proposed grid structure encodes kinematic forward relation (between parent and child) and peer relation (between peer joints) into a bidirectional grid coordinate system, which enables convolution operators to learn different motion contexts in respective directions.

\noindent\textbf{Feature mapping scheme}.
See Figure~\ref{fig:fcn_gcn_gridconv} for conceptual comparisons. 
FCN methods~\cite{martinez2017simple,li2020cascaded} connects all channels of the body joints to extract joint features with a learnable parameter $W\in R^{J\times C^{out} \times J \times C^{in}}$ in each layer.
Typical GCN methods \cite{zhao2019semgcn} and \cite{cai2019exploitingspatialtemporal} preserve skeleton topology by adopting graph convolution operators on joint feature learning in a two-step manner.
Step 1 embeds the latent joint feature from one dimension to another by sharing a learnable parameter $W^{embed}\in R^{C^{out}\times C^{in}}$ among joints.
Step 2 aggregates neighboring joint features to the center with a learnable parameter $W^{aggr}\in R^{J\times J}$, which does not involve changes in the size of feature channel.
It has limitation in the weight sharing scheme of $W^{emb}$, because each joint needs to aggregate neighboring joint features in a unique way, which has also been elucidated in \cite{ci2019lcn}.
GridConv implements single-step feature aggregation by defining standard convolution operation on grid structure.
Another GCN variant, LCN~\cite{ci2019lcn},  has also completed the reformulation from the two-step manner to the single-step one by proposing a locally connected operator.

In Table 1, 2, 3 of the main paper, we compare the performance of our method to FCNs and GCNs on two public benchmarks and results show significant margins of ours advanced to them.
Next, we further investigate where the superiority comes from in the ablation study "Grid vs. Graph" of the main paper, as shown in Table 4 of the main paper.
To train GCN methods on grid pose $D_{grid}$, we modify the convolution operator in SemGCN and LCN.
For SemGCN, input pose is replaced by grid pose with size of 5\texttimes5, and the aggregation operation of step 2 is modified to aggregate features from neighboring 8 nodes.
For LCN, input pose is replaced in the same way, and the original adjacency mask matrix is modified to connect each grid node to its 8 neighbors.
The results in column $D_{grid}$ indicate that single-step feature mapping scheme (LCN, GridConv) is more effective than two-step one (SemGCN) when migrating on grid representation. 
This is one factor that makes GridConv superior to traditional GCN.
Carefully designining two-branch architecture of GridConv module is another factor that helps learn pose lifting, which offers performance gain of GT input on Human3.6M benchmark from 40.3 (one branch, worse than LCN) to 39.0 (two branches, better than LCN).
We discuss the experiment in Section 3.5 of this document.
AutoGrids and Grid-specific attention modules constitute the third and the fourth factors, which enhance the performance of GridConv in a great leap forward and achieve the leadership against attention-based GCN method \cite{zeng2021skeletalgcn}.

\subsection{Jointly Training of 2D Detector and GLN}
We mainly follow the training scheme proposed in \cite{ci2020locally}.
In the experiment,
we use HRNet~\cite{sun2019hrnet} model as 2D detector, and modify the model by adding an upsampling layer to enlarge the heatmap output from 96\texttimes72 to 384\texttimes288 for more accurate keypoint detection.
The heatmap output in high resolution is delivered into a soft-argmax layer~\cite{sun2018integral} to generate 2D pose coordinate $G_{_{2D}}$.
Then grid lifting network takes $G_{_{2D}}$ as input and infers the 3D grid pose estimate $D_{_{3D}}$.
For loss function in the end-to-end learning process, we use 2D pose coordinate supervision and 3D pose coordinate supervision. 
The overall loss function $\mathcal{L}_{e2e}$ is defined as follows:
\begin{eqnarray}
	\mathcal{L}_{e2e} = \lVert  G_{_{2D}}  - G_{_{2D}}^{^{GT}} \rVert^{^1} + \lambda \lVert G_{_{3D}} - G_{_{3D}}^{^{GT}} \rVert^{^2},
\end{eqnarray}
where $\lambda$ is set to 160. 2D loss is measured by $L_1$ norm and 3D loss is measured by $L_2$ norm.

\subsection{Relationship with Existing Attention Methods}
In Section 3.3, attention mechanism is introduced to strengthen the learning ability of GridConv.
The grid-based attention module draws on the ideas of existing methods while also differs from them.
Existing attention methods in this line can be divided into three basic groups: (1) applying the attention mechanism conditioned on the input feature to re-calibrate the output feature of a convolutional layer, such as SENet~\cite{hu2018senet} and CBAM~\cite{woo2018cbam}; (2) applying the attention mechanism conditioned on the input feature to modify the weights of a convolutional layer, such as WeightNet \cite{ma2020weightnet} and CGC \cite{lin2020context}; (3) applying the attention mechanism over $n$ additive convolutional kernels to increase the size and the capacity of a network while maintaining efficient inference, such as CondConv~\cite{yang2019condconv} and DynamicConv~\cite{chen2020dynamic}. 

To these methods and our Dynamic Grid Convolution (D-GridConv), the structure of the attention function $\pi(\cdot)$ mostly follows the design of SENet, and their major difference lies in how the attention function is used. Although D-GridConv is not the key contribution of our paper, it differs from them both in focus and design. On the one side, D-GridConv aims to improve the ability of our GridConv to encode contextual cues (particularly to make GridConv operations spatial-aware and grid-specific). As we have discussed, GridConv acts as the basic building block of our grid lifting network, formulating a new representation learning paradigm for 2D to 3D human pose lifting task. Under this context, D-GridConv is also new and proven to be effective to improve the accuracy of our grid lifting network (see Table 9 of the main paper). On the other side, in the design of D-GridConv (see Equation 7 and Equation 8 of the main paper), the attention function $\pi(\cdot)$  generates a group of attention matrices $\{\alpha_{ij} \in \mathbb{R}^{K\times K} | i\in[1,H],\, j\in[1,P]\}$ with equal size to the grid, in which each $\alpha_{ij}$ corresponds to a $K$\texttimes$K$ grid patch. That is, for a grid pose pattern with the size of $H \times P$, D-GridConv applies $H$\texttimes$P$ different attentive scaling factor matrices $\alpha_{ij} \in \mathbb{R}^{K\times K}$ to adjust the shared grid convolution kernel $W$ grid by grid (i.e., centered grid), making grid convolution operations \textit{spatial-aware and grid-specific} (in other words, the dynamic grid convolution operation in the same layer is different but not shared to different grid patches of an input 2D pose sample). This is clearly different from the aforementioned dynamic/attentive convolution methods (in image space) where attentive scaling factors are usually shared to the whole feature channel or all pixels or the whole convolutional kernel.

\section{More Ablative Experiments}
\subsection{Comparison in Protocol 2}
In Table~\ref{tab:pa-mpjpe}, we compare our method with related lifting works under Protocol 2. 
Protocol 2 uses 2D detection as input and transforms 3D prediction to the ground truth with rigid alignment. 
Our model under two settings (standard normalization and the one marked as §) can achieve state-of-the-art accuracy in comparison to existing methods including the GCNs~\cite{ci2019lcn,cai2019exploitingspatialtemporal,zeng2021skeletalgcn}.
Meanwhile, the performance superiority of setting § in comparison to standard normalization setting under Protocol 1 (46.3 \textit{vs} 47.6) disappears when measuring under Protocol 2.

\subsection{More Variants of Handcrafted Grid Pose}
One may concern about the effect of other handcrafted grid poses. We performed a set of experiments to explore this problem. Specifically, we take the well-designed handcrafted grid pose as the reference and apply three types of grid shuffle methods over it to generate more handcrafted grid pose variants for comparison.
From the results shown in Table~\ref{tab:shuff_standard}, we can see that all shuffle patterns perform worse than the handcrafted reference, and 
Column Shuffle works better than Row Shuffle and Global Shuffle. It is because peer joints implied in each column do not have definite order. Hence the skeleton topology is still preserved in column shuffled grid pose patterns.

\begin{table}\small
	\centering
	\begin{tabular}{l|ccc}
		\toprule
		\multirow{2}{*}{Grid Shuffle Method}  & \multicolumn{3}{c}{MPJPE}     \\ 
		& P1   & P1*  & P2   \\ \midrule
		No Shuffle  & \textbf{47.9} & \textbf{37.1} & \textbf{37.9} \\
		Row Shuffle  & 49.2 & 38.4 & 38.2     \\
		Column Shuffle & 48.4 & 37.7 & 38.1   \\
		Global Shuffle & 49.9 & 39.4 & 38.4 \\ \bottomrule
	\end{tabular}
	\caption{Ablation study on changing handcrafted grid pose layout by different shuffle methods. Row Shuffle changes the order of parent and child grid nodes. Column Shuffle changes the order of peer grid nodes. Global Shuffle rearranges all grid nodes.}
	\label{tab:shuff_standard}
\end{table}

\subsection{Temporal GridConv}
One intuitive question is whether GridConv can receive temporal input, which tests the validity of GridConv in a broad sense.
To figure it out, we implement temporal 3D grid convolution based on VPose~\cite{pavllo2019videopose}.
The temporal network perceives 243-frame human poses by stacking four residual blocks, each of which has a 1D convolution module of kernel size $K=3$ and a 1D convolution module of $K=1$.
We replace the first convolution module by 3D D-GridConv of temporal kernel size $K_t=3$ and spatial kernel size $K_s=3$, and replace the second one by 3D D-GridConv of $K_t=1$ and $K_s=3$.
The padding size of spatial dimension is set to 1 on each side to preserve the spatial size of the pose sequence at $H$\texttimes$P$ throughout network layers.
The rest settings keep the same as the original framework.
So far we have only tried handcrafted SGT in this experiment since integrating learnable SGT into the framework requires a more careful training design.
Due to the longer training time, we have tried temporal input only with 27 frames via stacking two residual blocks in the temporal network.
As shown in Table~\ref{tab:video}, the results of two input sources show that our temporal D-GridConv works well and has better accuracy than the baseline VideoPose-243f with only given 27-frame input.
These results demonstrate that the proposed grid convolution is effective on temporal input as well. But there are still many factors under study, such as efficiency, integration of learnable SGT, the possibility of reaching SOTA accuracy, and generalization ability on new scenes or different FPS. 
The scope is beyond the focus of this paper, so we leave it as our future work.

\begin{table}[]\small
\centering
\begin{subtable}[]{\linewidth}
	\centering
	\begin{tabular}{l|r|cc}
		\toprule
		\multirow{2}{*}{Method} & \multirow{2}{*}{Parameters} & \multicolumn{2}{c}{MPJPE} \\
		&                            & P1*         & P2*                       \\ \midrule
		VPose\shortcite{pavllo2019videopose} 27f  & 8.6 M      & 40.6  & -       \\
		VPose\shortcite{pavllo2019videopose} 243f & 17.0 M     & 37.2  & 27.2    \\
		Ours 27f       & 38.1 M     & \textbf{33.5}  & \textbf{24.7}   \\ \bottomrule
	\end{tabular}
	\caption{Ground truth 2D as input.}
\end{subtable}
\\
\begin{subtable}[t]{\linewidth}
	\centering
	\begin{tabular}{l|r|cc}
		\toprule
		\multirow{2}{*}{Method} & \multirow{2}{*}{Parameters} & \multicolumn{2}{c}{MPJPE} \\
		&                            & P1         & P2                    \\ \midrule
		VPose\shortcite{pavllo2019videopose} 243f & 17.0 M     & 46.8  & 36.5    \\
		Ours 27f       & 38.1 M     & \textbf{44.3}  & \textbf{33.2}   \\ \bottomrule
	\end{tabular}
	\caption{Detection results of CPN as input.}
\end{subtable}
\caption{Performance comparison of VPose and Temporal GridConv given 2D pose sequence as input.}
\label{tab:video}
\end{table}

\subsection{Multi-branch Architecture}
Two-branch architecture is a basic module design of GridConv as shown in the network overview of the main paper.
Here we dig deep into the impact of multi-branch architecture on estimation accuracy.
Experiments include two parts, one-branch architecture (using two padding modes) and multi-branch architecture (composition of the one-branch).
Results can be found in Table~\ref{tab:padding}.

Table~\ref{subtab:one-branch} and \ref{subtab:two-branch} show that multi-branch architectures have remarkable gains compared to any of the one-branch one. 
This finding makes multi-branch architecture one of the core design of grid convolution, although it is not emphasized in the main paper.
When the number of branches is increased to four, the gain is very small compared to the best performance of two branches. 
It means that the two-branch architecture tends to be saturated. 
For efficiency reasons, two-branch architecture \#1+\#2 is set as the default for GridConv module and D-GridConv one.

\begin{table}[tb]\small
\centering
\begin{subtable}[t]{\linewidth}\small
	\centering
	\begin{tabular}{l|cc|c}
		\toprule
		\multirow{2}{*}{Name} & \multicolumn{2}{c|}{Padding Mode} & \multirow{2}{*}{MPJPE} \\ 
		& Horizontal       & Vertical       &                        \\ \midrule
		\#1                       & \textit{Circular}             & \textit{Circular}            & 40.8                   \\
		\#2                       &\textit{Replicate}             & \textit{Replicate}           & 42.3                   \\
		\#3                      & \textit{Circular}              & \textit{Replicate}           & 41.1                   \\
		\#4                        & \textit{Replicate}             & \textit{Circular}            & 40.4                   \\ \bottomrule
	\end{tabular}
	\caption{One-branch architecture inside vanilla GridConv module.}
	\label{subtab:one-branch}
	\vspace{0.5em}
\end{subtable}	

\begin{subtable}[tb]{\linewidth}\small
	\centering
	\begin{tabular}{ccccc}
		\toprule
		\multicolumn{4}{c|}{Composition}                                                            & \multirow{2}{*}{MPJPE} \\ 
		\multicolumn{1}{c}{\#1} & \multicolumn{1}{c}{\#2} & \multicolumn{1}{c}{\#3} & \multicolumn{1}{c|}{\#4} &                        \\ \midrule
		\multicolumn{5}{c}{Two Branches} \\ \midrule
		\multicolumn{1}{c}{$\times2$}& \multicolumn{1}{c}{-}      & \multicolumn{1}{c}{-}       &  \multicolumn{1}{c|}{-}    & 40.2                       \\
		\multicolumn{1}{c}{-} & \multicolumn{1}{c}{$\times2$}& \multicolumn{1}{c}{-}       &  \multicolumn{1}{c|}{-}    & 41.2                       \\
		\multicolumn{1}{c}{-}      & \multicolumn{1}{c}{-}       & \multicolumn{1}{c}{$\times2$}       &   \multicolumn{1}{c|}{-}   &     41.5                   \\
		\multicolumn{1}{c}{-}      & \multicolumn{1}{c}{-}       & \multicolumn{1}{c}{-}       &   \multicolumn{1}{c|}{$\times2$}   &  40.5                      \\
		\multicolumn{1}{c}{\checkmark}      & \multicolumn{1}{c}{\checkmark}       & \multicolumn{1}{c}{-}       &   \multicolumn{1}{c|}{-}   &  \textbf{39.0}                      \\
		\multicolumn{1}{c}{\checkmark}      & \multicolumn{1}{c}{-}       & \multicolumn{1}{c}{\checkmark}       &   \multicolumn{1}{c|}{-}   &   39.2                     \\
		\multicolumn{1}{c}{\checkmark}      & \multicolumn{1}{c}{-}       & \multicolumn{1}{c}{-}       &   \multicolumn{1}{c|}{\checkmark}   &   39.3                     \\
		\multicolumn{1}{c}{-}      & \multicolumn{1}{c}{\checkmark}       & \multicolumn{1}{c}{\checkmark}       &  \multicolumn{1}{c|}{-}    &    40.6                    \\
		\multicolumn{1}{c}{-}      & \multicolumn{1}{c}{\checkmark}       & \multicolumn{1}{c}{-}       &  \multicolumn{1}{c|}{\checkmark}    &  39.7                      \\
		\multicolumn{1}{c}{-}      & \multicolumn{1}{c}{-}       & \multicolumn{1}{c}{\checkmark}       &  \multicolumn{1}{c|}{\checkmark}    &  39.5                      \\ \midrule
		\multicolumn{5}{c}{Four Branches} \\ \midrule
		\multicolumn{1}{c}{\checkmark}      & \multicolumn{1}{c}{\checkmark}       & \multicolumn{1}{c}{\checkmark}       &  \multicolumn{1}{c|}{\checkmark}    &  \textbf{38.9}                      \\
		\bottomrule
	\end{tabular}
	\caption{Multi-branch architecture inside vanilla GridConv module.}
	\label{subtab:two-branch}
\end{subtable}
\caption{Comparison of GridConv modular architecture with different padding modes. Multi-branch architectures are trained using GT 2D input.}
\label{tab:padding}
\vspace{-1.5em}
\end{table}

\subsection{Convolutional Kernel Pattern}
In the main paper, we fix convolutional kernel size to $K=3$ for all D-GridConv modules.
Here we provide an ablative experiment on the effect of different kernel sizes to validate our choice.

We tried three convolutional kernel sizes at 1/3/5 and designed four combination types of them in D-GridConv layers.
We can see the settings in Table~\ref{tab:kernel_size}. 
All combination types obey the \textit{(a-bc-bc-d)} form.
Specifically, the network is composed of an expanding D-GridConv layer with kernel size \textit{a}, two D-GridConv layers in the residual blocks with kernel size \textit{b} and \textit{c}, and a shrinking D-GridConv layer with kernel size \textit{d}.
As the size of grid pose is set at 5\texttimes5, the convolution layer with a 5\texttimes5 kernel aggregates feature of all grids at one time.
The one with a 3\texttimes3 kernel aggregates features of 8 neighbors to the central grid in a local patch.
The one with a 1\texttimes1 kernel only embeds features of each grid to another dimension without aggregating neighboring information.

The results demonstrate that all combination types present good performance under three protocols, which indicates that our grid lifting model is not sensitive to the setting of kernel size.
Although all combinations can produce sound 3D estimates, type \#4 using $K=3$ in all layers outstands from the rest on the performance with noticeable margins.
We take it as the default setting of the convolutional layer design to pursue the best performance of the grid lifting network.

\begin{table}[]\small
\centering
\begin{tabular}{l|c|ccc}
	\toprule
	\multirow{2}{*}{Kernel Size} & \multirow{2}{*}{Parameters \#} & \multicolumn{3}{c}{MPJPE} \\ 
	& &  P1          & P1*   & P2      \\ \midrule
	Type \#1 (3-33-3)           &   4.79M     & \textbf{47.9} & \textbf{37.1} & \textbf{37.9}   \\ 
	Type \#2 (3-31-3)           &    2.69M &   48.3  &   37.2    & 38.0 \\
	Type \#3 (3-53-3)           &    9.00M &  48.9   &    37.8     & 37.9\\
	Type \#4 (3-33-1)           &    4.77M &   49.8  &    37.7    & 38.1\\    
	\bottomrule
\end{tabular}
\caption{Performance comparison on different combinations of kernel size. ($a$-$bc$-$d$) means kernel size $K$=$a$ for the first dimension expanding layer, $K$=$b$ for the first layer, and $K$=$c$ for the second layer in residual blocks, and $K$=$d$ for the last dimension shrinking layer.}
\label{tab:kernel_size}
\end{table}

\subsection{Impact on Model Size}
\begin{figure}
\begin{subfigure}{0.48\linewidth}
	\centering
	\includegraphics[width=\linewidth]{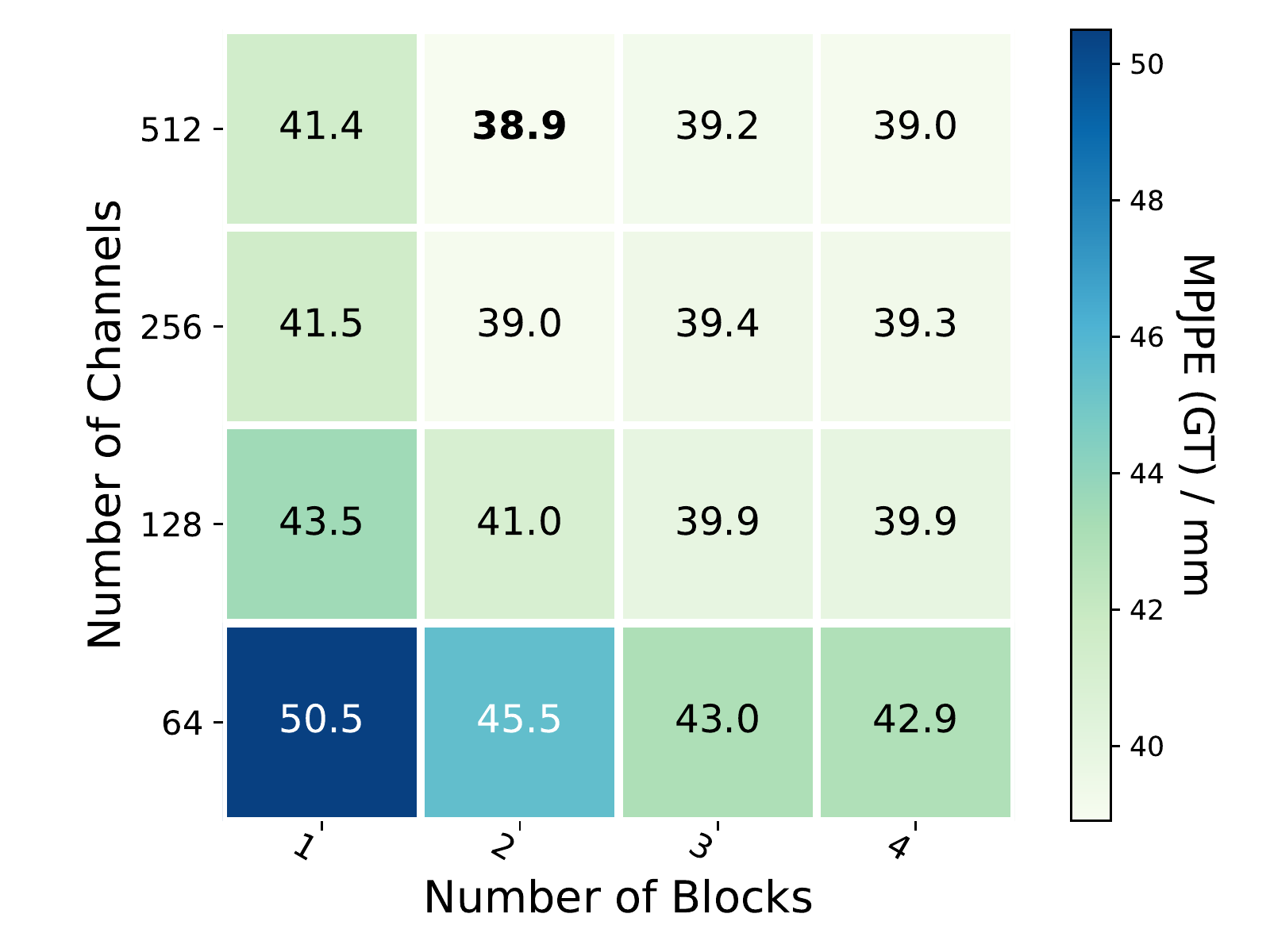}
	\caption{Accuracy.}
	\label{fig:acc}
\end{subfigure}
\begin{subfigure}{0.48\linewidth}
	\centering
	\includegraphics[width=\linewidth]{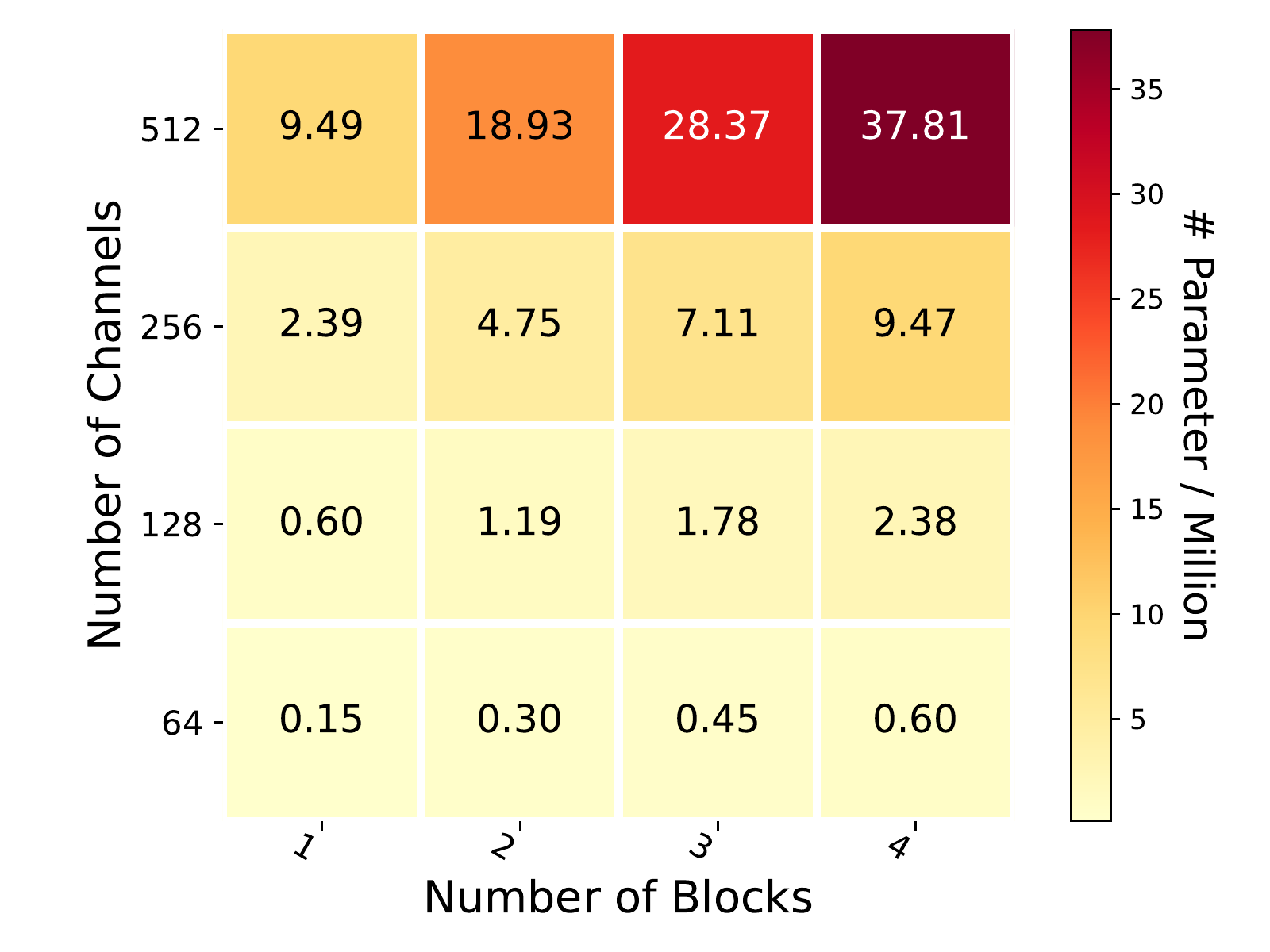}
	\caption{Model size.}
	\label{fig:param}
\end{subfigure}
\caption{Performance of our models with different numbers of residual blocks and latent channels.
	Models are trained under GT 2D input.}
\label{fig:depth_width}
\vspace{-1.5em}
\end{figure}

A sufficiently good GLN baseline is critical for the proposed designs.
To this end, we investigate the number of residual blocks $B$, and the number of channels of latent grid pose $C$.
We use the handcrafted model with vanilla GridConv modules in this experiment.
The results are shown in Figure~\ref{fig:depth_width}.

Figure~\ref{fig:acc} shows that accuracy tends to be saturated when the number of residual blocks goes to 3 for either narrow or wide networks.
When the channel number of latent features reaches $C=256$, the improvement from continually widening the network tends to be very small.
The lowest MPJPE=$38.9mm$ is achieved by the model of $B=2, C=512$.
However, in this case, the number of trainable model parameters reaches up to $18.93M$, even larger than a few temporal models, which is intolerable.
For the accuracy-efficiency tradeoff, we thus fix $B=2, C=256$ as the default setting.

\begin{figure}[h]
	\centering
	\begin{subfigure}{0.7\linewidth}
		\centering
		\includegraphics[width=\linewidth]{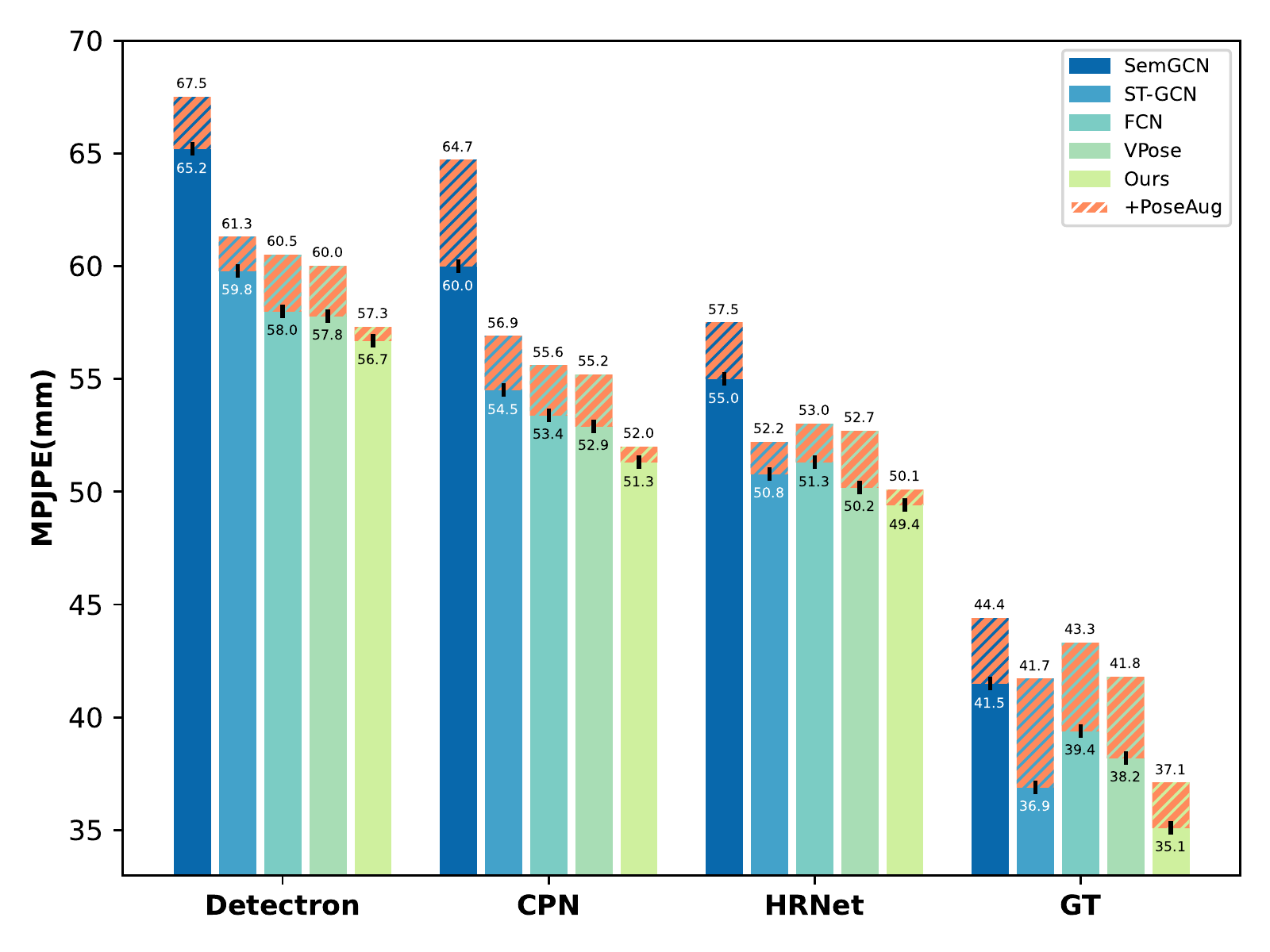}
		\caption{Performance on Human3.6M.}
		\label{subfig:aug_h36m}
	\end{subfigure}
	\begin{subfigure}{0.7\linewidth}
		\centering
		\includegraphics[width=\linewidth]{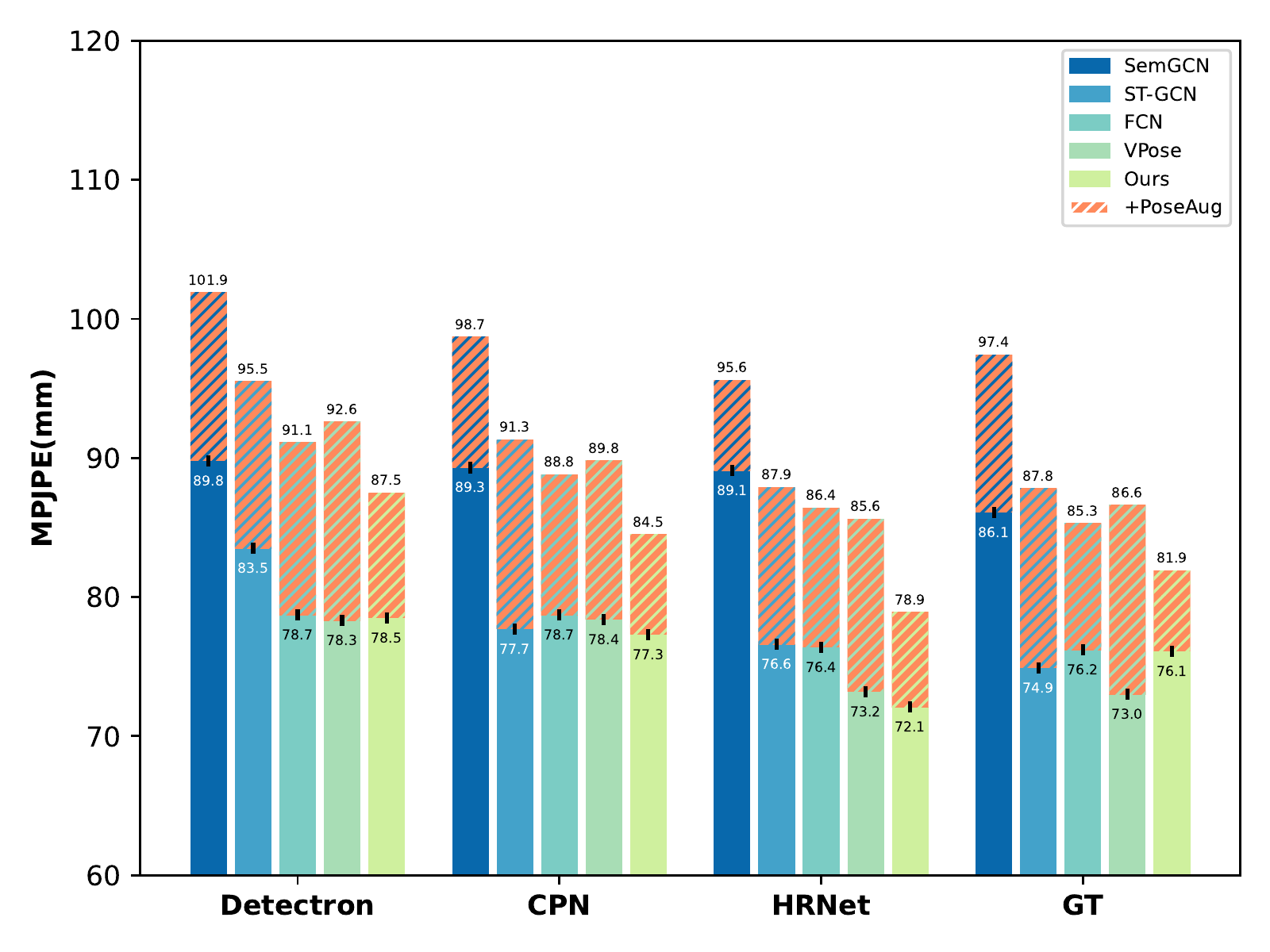}
		\caption{Performance on MPI-INF-3DHP.}
		\label{subfig:aug_3dhp}
	\end{subfigure}
	\caption{Performance comparison when training with synthetic data. Indicators above orange bars are baseline results and those under the bars are augmented training results.}
	\label{fig:aug}
	\vspace{-1em}
\end{figure}

\begin{figure}
	\centering
	\begin{subfigure}{0.4\linewidth}
		\centering
		\includegraphics[width=\linewidth]{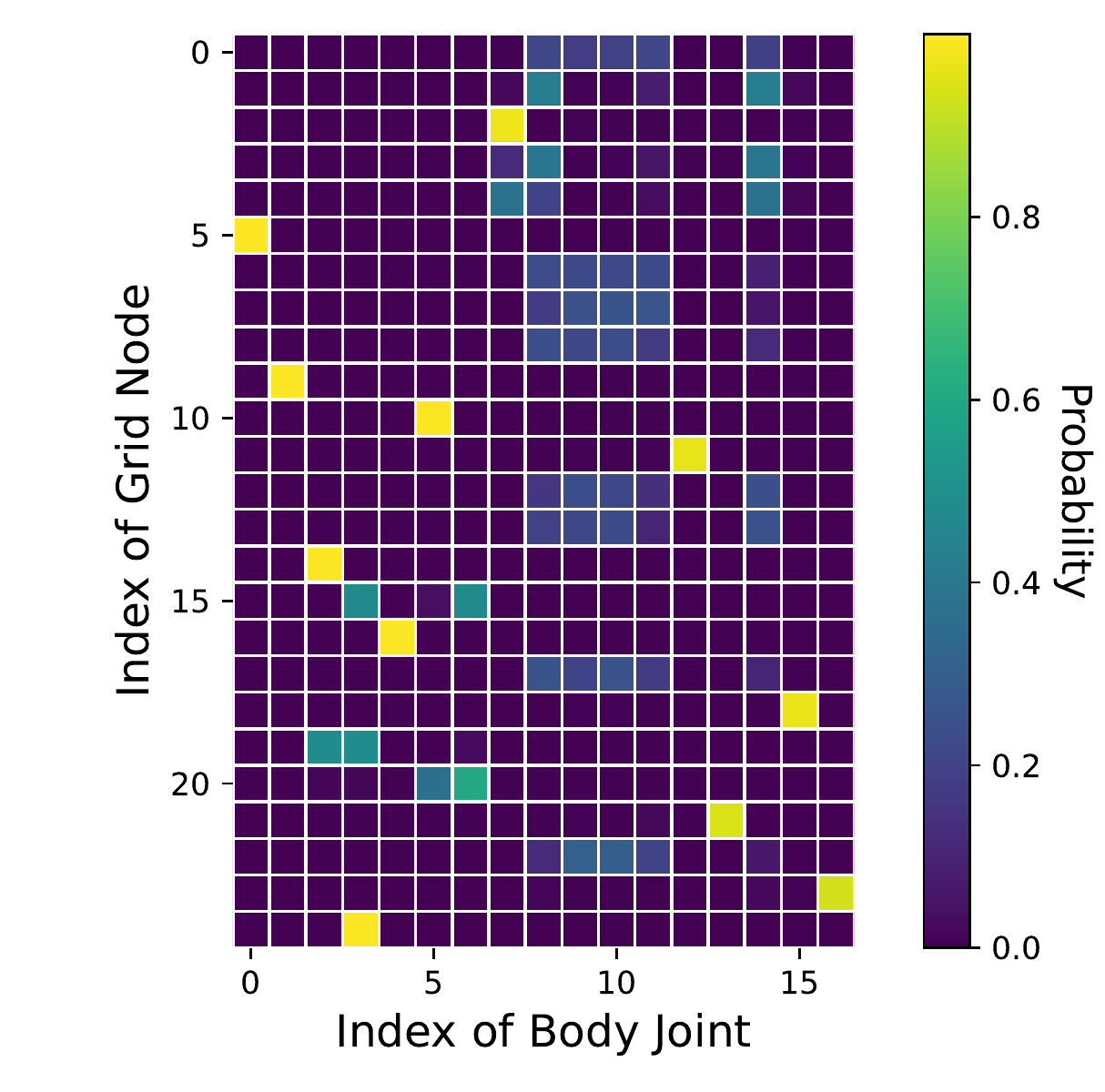}
		\caption{Probability}
		\label{fig:prob}
	\end{subfigure}
	\begin{subfigure}{0.4\linewidth}
		\centering
		\includegraphics[width=\linewidth]{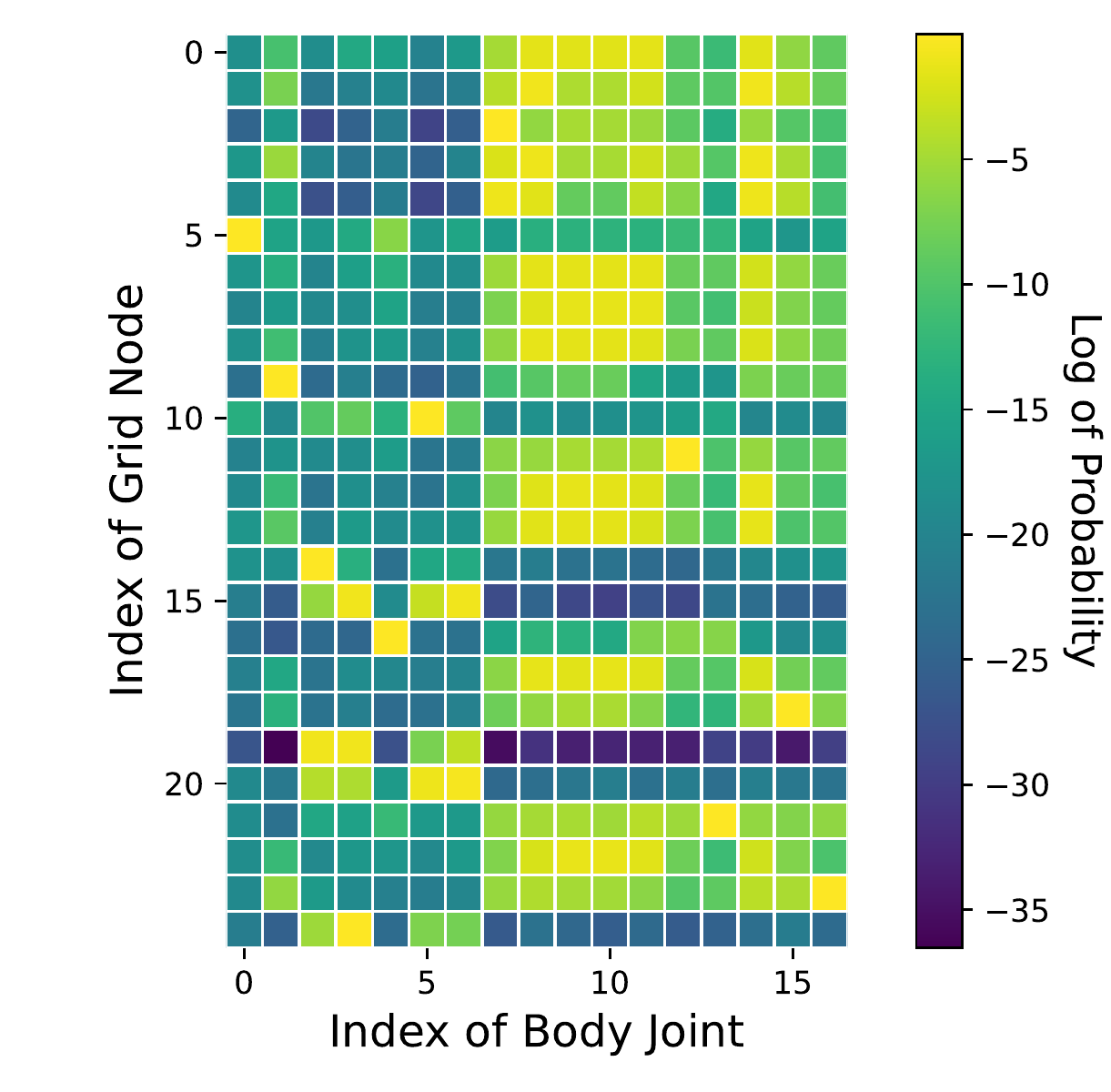}
		\caption{Log of probability.}
		\label{fig:log_prob}
	\end{subfigure}
	\begin{subfigure}{0.18\linewidth}
		\centering
		\includegraphics[height=2.45\linewidth]{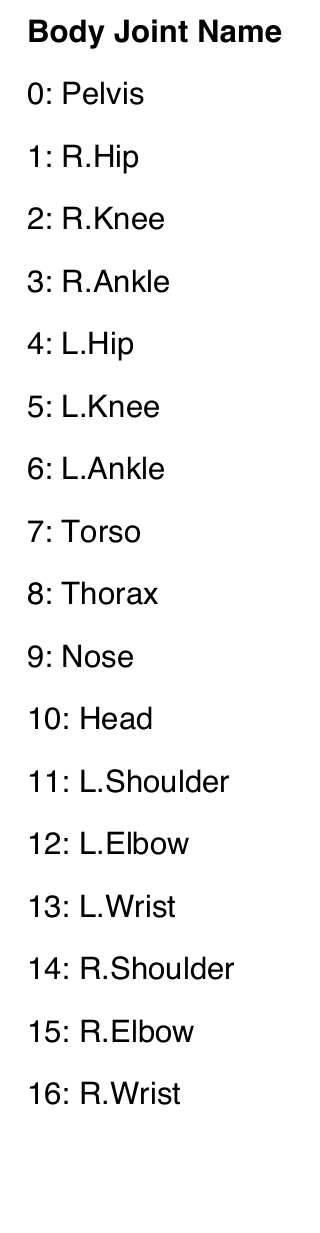}
	\end{subfigure}
	\caption{Visualization on assignment probability matrix $S^{prob}$.}
	\label{fig:prob_mat_vis}
	
\end{figure}

\subsection{Boosting Performance by Augmented Training}
As we mentioned in Related Work section, several works~\cite{li2020cascaded,gong2021poseaug,fang2018posegrammar} proposed sophisticated startegies to generate synthetic 2D-to-3D data for fine-tuning FCN or GCN lifting models.
To investigate the learning potential of our grid lifting network under the augmented training condition, we use PoseAug~\cite{gong2021poseaug} framework to train our Handcrafted SGT model with synthetic data.
The framework generates 16-joint samples by discarding the neck joint.
To ensure our model run properly, we interpolate the neck by averaging the values of left and right shoulder joints.
MPJPE is still measured on the 16 joints for all methods.
For more complete comparison, the experiment sets up four input sources: Detectron~\cite{Detectron2018}, CPN~\cite{chen2018cpn}, HRNet~\cite{sun2019hrnet}, and ground truth.

We compare our results with four existing works~\cite{martinez2017simple,zhao2019semgcn,cai2019exploitingspatialtemporal,pavllo2019videopose}, as shown in Figure~\ref{fig:aug}. 
Among the baseline results, our model outperforms other methods by remarkable margins on Human3.6M (Figure \ref{subfig:aug_h36m}) and MPI-INF-3DHP (Figure \ref{subfig:aug_3dhp}).
When training with PoseAug, our model gets improvements from augmented data, still surpasses all the rest on Human3.6M, and achieves comparable performance under cross-dataset evaluation on MPI-INF-3DHP.

\subsection{Visualization of Learnt Assignment Matrix}
As we mentioned in the main paper, AutoGrids learns a probability distribution for assignment matrix $S$ from body joints to grid nodes.
In Figure~\ref{fig:prob_mat_vis}, we visualize the probability matrix of learned grid pattern \#2 depicted in Figure 7c of the main paper.
In Figure~\ref{fig:prob}, there are some rows that have high-probability cells visualized as yellow.
A yellow cell at position $(i,j)$ indicates that the body joint $i$ is assigned to the corresponding grid $j$ with high confidence.
While the rest rows having no high-responded node tend to learn two to four proposals in similar possibilities.
In Figure~\ref{fig:log_prob}, the log heatmap indicates that the learner has a higher focus on joints 7-16, which build up the upper body.
This is probably because upper body movements in the selected benchmark, Human3.6M, appear more diverse than the lower body ones. 
Focusing on these information-rich joints helps pose feature modeling.

\subsection{Visualization of Convolutional Kernel Activations}
In the main paper, we formulate D-GridConv as a dynamic variant of vanilla GridConv by making it grid-specific, spatial-aware, and input-dependent.
Here we provide a comprehensive analysis of the convolutional kernel of D-GridConv, including three aspects: across channels (or filters), across spatial locations, and across input samples. 
All visualizations are conducted on the handcrafted SGT model given ground-truth input.

\begin{figure}[t]
	\centering
	\begin{subfigure}{\linewidth}
		\centering
		\includegraphics[width=0.66\linewidth]{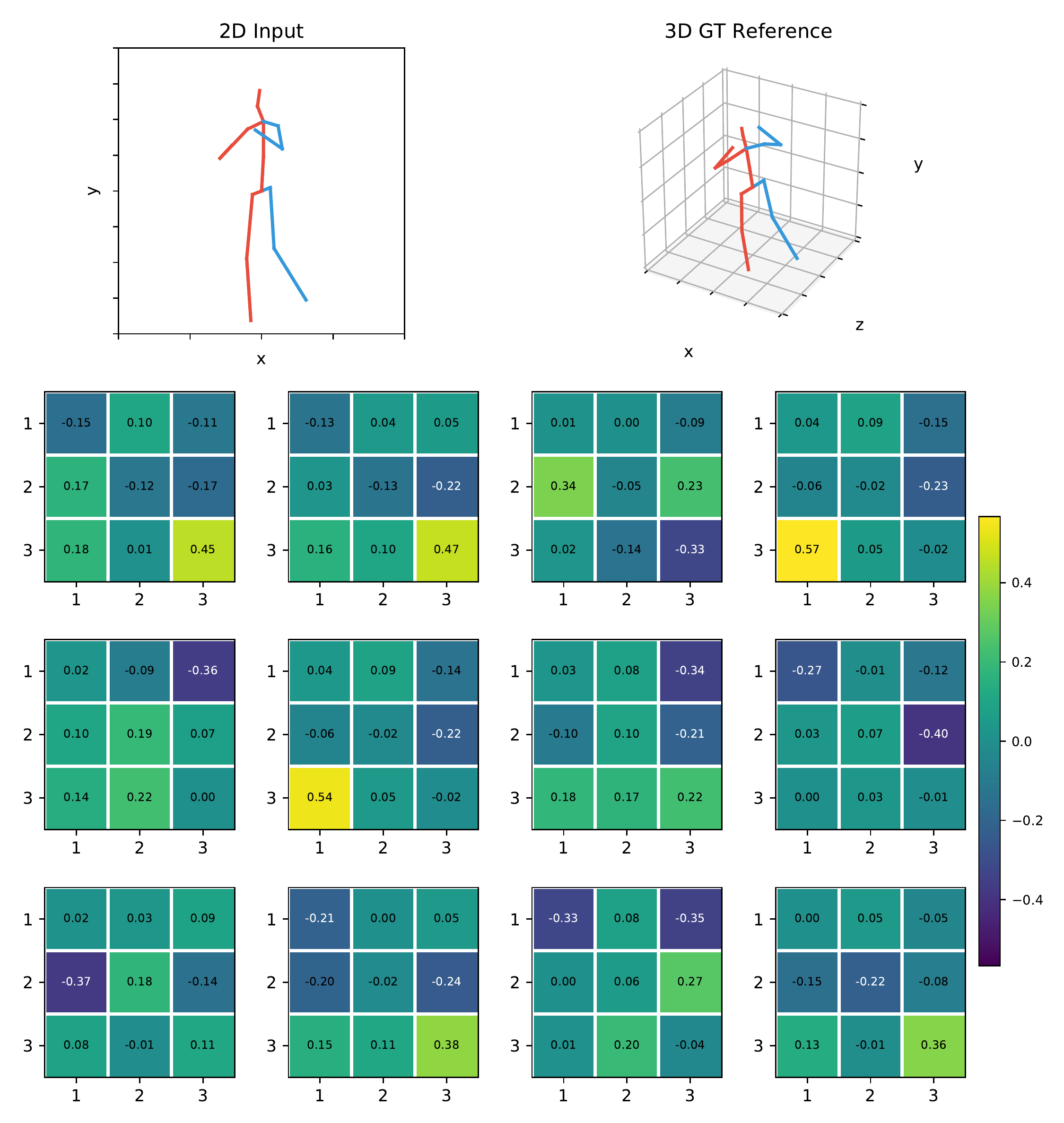}
		\caption{Given input sample \#1.}
		\label{fig:sample1}
		\vspace{1ex}
	\end{subfigure}

	\begin{subfigure}{\linewidth}
		\centering
		\includegraphics[width=0.66\linewidth]{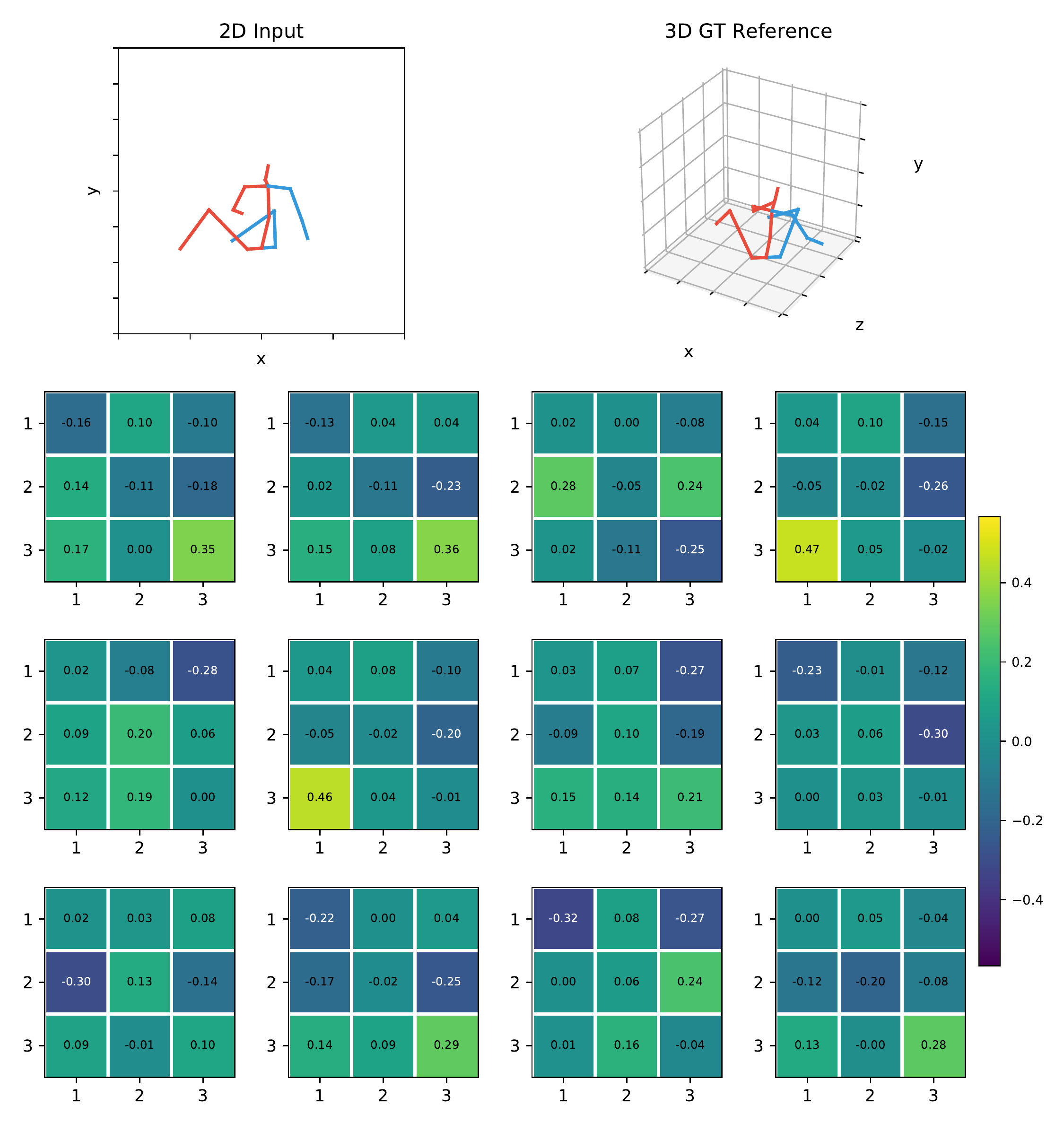}
		\caption{Given input sample \#2.}
		\label{fig:sample2}
	\end{subfigure}
	\caption{Visualization on D-GridConv kernel dynamically changing according to the input samples.}
	\label{fig:kernel_sample}
\end{figure}

\paragraph{Across input samples}
In Figure~\ref{fig:kernel_sample}, we randomly visualize several D-GridConv kernels, in which the weights vary according to the input sample.
The input sample \#1 belongs to action \textit{Directions} and the sample \#2 belongs to action \textit{SittingDown}.
The weight difference when giving different input samples has a similar trend to that when giving different spatial locations, which tends to adjust large weights.

\begin{figure}[t]
	\centering
	\includegraphics[width=\linewidth]{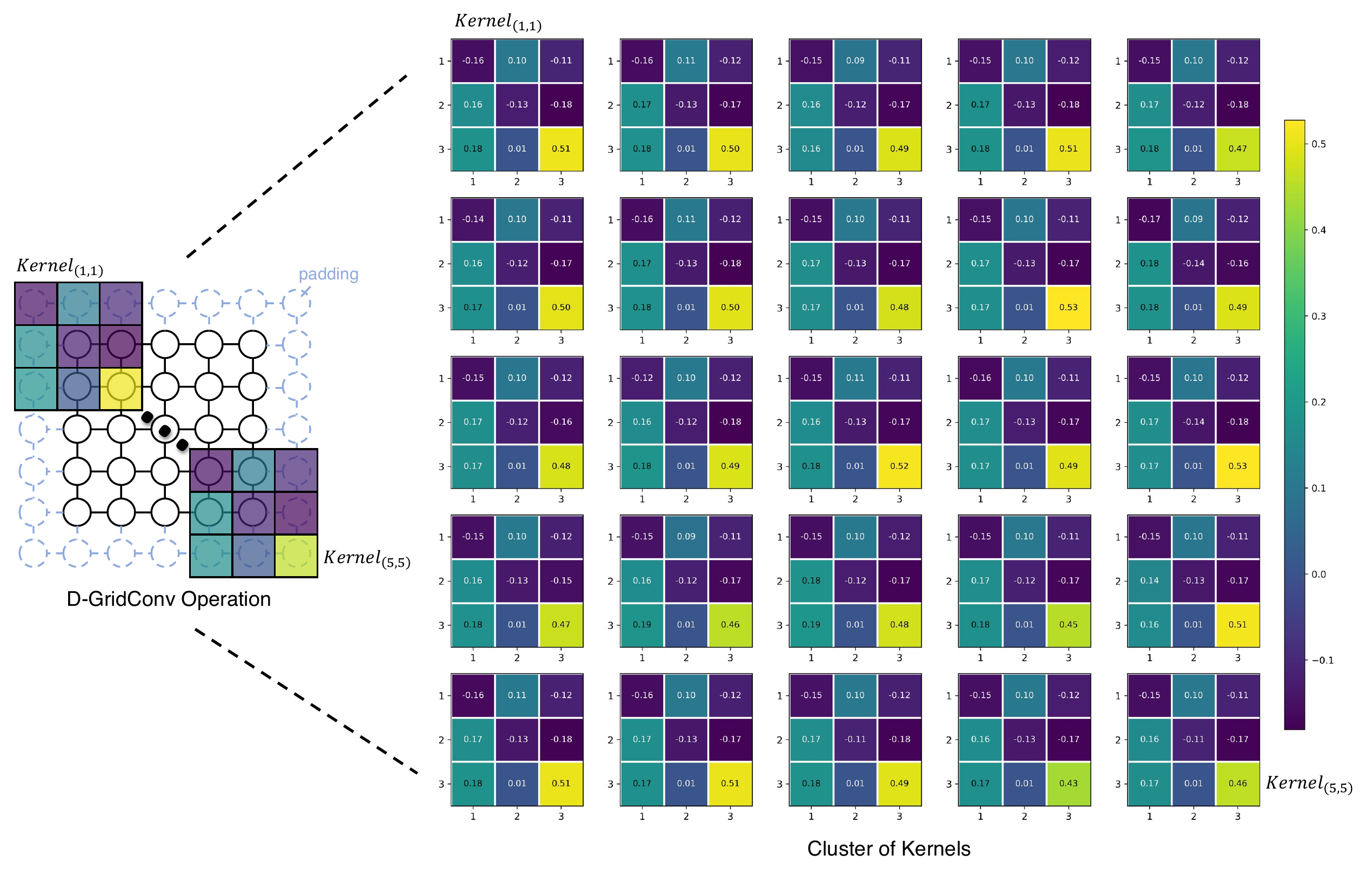}
	\vspace{-1em}
	\caption{Visualization on D-GridConv kernel dynamically changing according to spatial locations.}
	\label{fig:w1_5x5}
\end{figure}

\paragraph{Across spatial locations}
In Figure~\ref{fig:w1_5x5}, we visualize the most-varied channel of the D-GridConv kernels, which varies according to spatial locations in the expanding layer.
The difference mainly appears in the lower right corner, which has the largest magnitude and the largest impact on the output.
This indicates that the attention module does not drastically change all of the kernel weights, but rather adjusts weight magnitude in small increments.

\begin{figure}
	\centering
	\begin{subfigure}{0.19\linewidth}
		\centering
		\rotatebox{90}{\includegraphics[width=360pt]{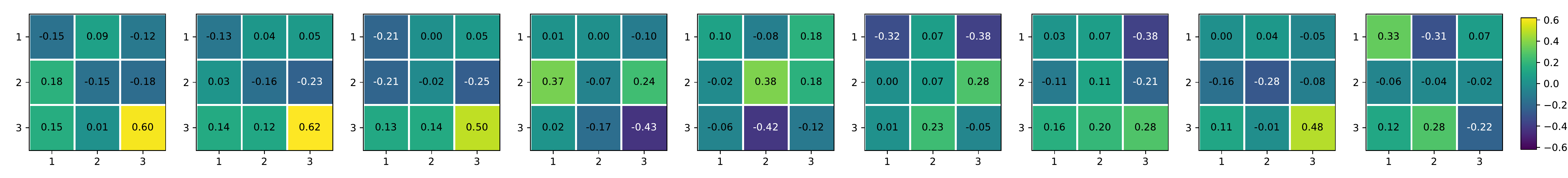}}
		\caption{U dim.}
		\label{fig:w1_x}
	\end{subfigure}
	\begin{subfigure}{0.19\linewidth}
		\centering
		\rotatebox{90}{\includegraphics[width=360pt]{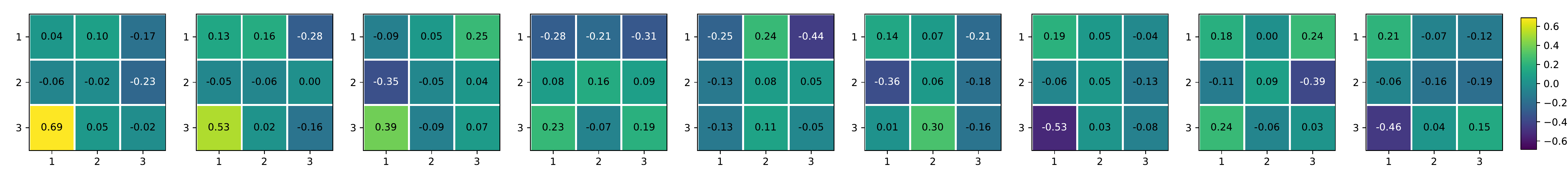}}
		\caption{V dim.}
		\label{fig:w1_y}
	\end{subfigure}
	\begin{subfigure}{0.19\linewidth}
		\centering
		\rotatebox{90}{\includegraphics[width=360pt]{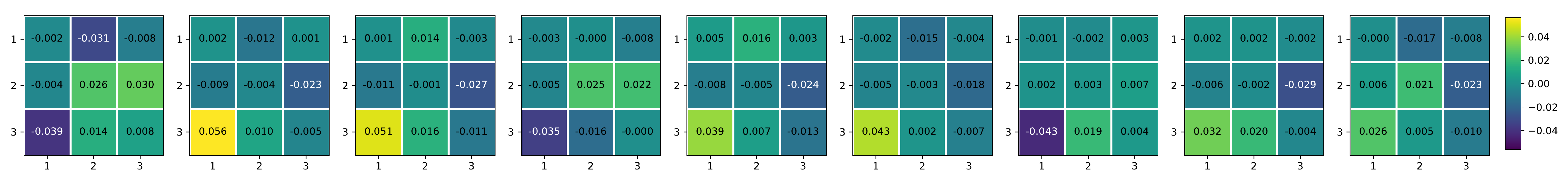}}
		\caption{X dim.}
		\label{fig:w2_x}
	\end{subfigure}
	\begin{subfigure}{0.19\linewidth}
		\centering
		\rotatebox{90}{\includegraphics[width=360pt]{figures/vis_kernel/dgridconv_w2_chn9_y.pdf}}
		\caption{Y dim.}
		\label{fig:w2_y}
	\end{subfigure}
	\begin{subfigure}{0.19\linewidth}
		\centering
		\rotatebox{90}{\includegraphics[width=360pt]{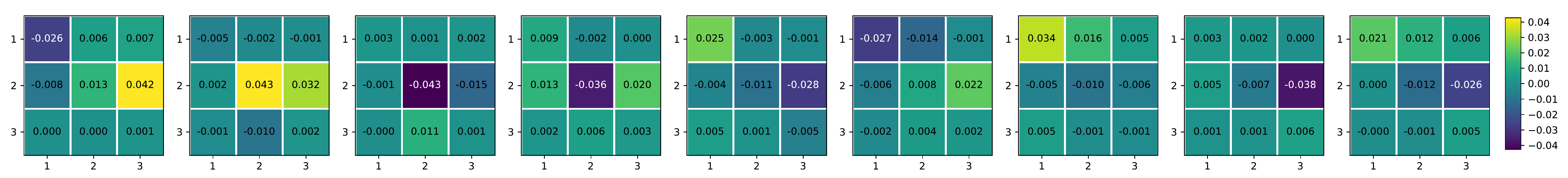}}
		\caption{Z dim.}
		\label{fig:w2_z}
	\end{subfigure}
	\vspace{-0.5em}
	\caption{Visualization on D-GridConv kernel in the first expanding layer for (a), (b), and in the last shrinking layer for (c), (d), (e). Select  most varied top nine channel items.}
	\label{fig:kernel_w}
	
\end{figure}

\paragraph{Across channels/filters}
In Figure~\ref{fig:kernel_w}, we visualize some D-GridConv kernels in the first expanding D-GridConv layer and the last shrinking D-GridConv layer respectively.
In those two figures, we can see that the kernel weights of different input channels/filters differ a lot.
On the other hand, the kernels are generally sparse with few large weights.
Large weights usually appear once in nine cells, such as the highlighted yellow ones in Figure~\ref{fig:w1_x} and \ref{fig:w1_y}.
As the magnitude variance continues to decrease, relatively large weights tend to appear more times.
However, in either case, small weights close to zero are in the majority, indicating that the D-GridConv kernels are generally sparse and focus on certain specific positions.

Overall, the visualization results in the above aspects illustrate how the attention module makes grid convolution kernels adaptive to spatial locations and input motions in order to suit our grid pose representation.

\end{document}